%% file: arxiv_main.tex
\documentclass[runningheads]{llncs}
\usepackage{graphicx}
\usepackage{amsmath,amssymb} % define this before the line numbering.
\usepackage{color}
\usepackage{isomath}
\usepackage[T1]{fontenc}    % use 8-bit T1 fonts
\usepackage{url}            % simple URL typesetting
\usepackage{booktabs}       % professional-quality tables
\usepackage{amsfonts}       % blackboard math symbols
\usepackage{nicefrac}       % compact symbols for 1/2, etc.
\usepackage{microtype}      % microtypography
\usepackage{xcolor}
\usepackage{epsfig}
\usepackage{multirow}
\usepackage{bbm}
\usepackage{bm}
\usepackage{textcomp}
\usepackage{footnote}
\usepackage{tablefootnote}

\usepackage[linesnumbered,ruled]{algorithm2e}
\usepackage{mathtools}
\usepackage{sidecap}
\usepackage{hyperref}
\usepackage{symbols}
\usepackage[export]{adjustbox}


\begin{document}
\title{Diverse and Coherent Paragraph Generation from Images} 
% Replace with your title

\titlerunning{Diverse and Coherent Paragraph Generation from Images}
% Replace with a meaningful short version of your title
%
\author{Moitreya Chatterjee \and
Alexander G. Schwing }
%
%Please write out author names in full in the paper, i.e. full given and family names. 
%If any authors have names that can be parsed into FirstName LastName in multiple ways, please include the correct parsing, in a comment to the volume editors:
%\index{Lastnames, Firstnames}
%(Do not uncomment it, because you may introduce extra index items if you do that, we will use scripts for introducing index entries...)
\authorrunning{Moitreya Chatterjee and Alexander G. Schwing}
% Replace with shorter version of the author list. If there are more authors than fits a line, please use A. Author et al.
%

\institute{University of Illinois at Urbana-Champaign, Urbana IL 61801, USA\\ 
\email{metro.smiles@gmail.com}, \email{aschwing@illinois.edu}}
\maketitle              % typeset the header of the contribution
\begin{abstract}
Paragraph generation from images, which has gained popularity recently, is an important task for video summarization, editing, and support of the disabled. Traditional image captioning methods fall short on this front, since they aren't designed to generate long informative descriptions. Moreover, the vanilla approach of simply concatenating multiple short sentences, possibly synthesized from a classical image captioning system, doesn't embrace the intricacies of paragraphs: coherent sentences, globally consistent structure, and diversity. To address those challenges, we propose to augment paragraph generation techniques with ``coherence vectors,'' ``global topic vectors,'' and modeling of the inherent ambiguity of associating paragraphs with images, via a variational auto-encoder formulation. We demonstrate the effectiveness of the developed approach on two datasets, outperforming existing state-of-the-art techniques on both. %by {\color{blue}xxx}\% and {\color{blue}xxx}\% respectively.

\end{abstract}

\input{introduction}
\input{related}
\input{method}
\input{evaluation}
%\input{analysis}
\input{conclusion}

\bibliographystyle{splncs04}
\bibliography{arXiv_main}

\clearpage
{\noindent \Large \textbf{Supplementary Material: Diverse and Coherent Paragraph Generation from Images}}

In this supplementary material, we first present examples of generated paragraphs from the Stanford Image-Paragraph dataset~\cite{krause2017hierarchical}, followed by some examples from the `Office-Products' category of the Amazon Product-Review Dataset~\cite{mcauley2015image}.  For completeness, we also show a derivation of the solution to the optimization problem of the \textit{Coupling Unit}.

%\vspace{0.8cm}

\section{Visualization of Generated Paragraphs}
%\vspace{0.3cm}

In this section, we first present sample paragraphs generated using images from the Stanford Image-Paragraph dataset~\cite{krause2017hierarchical}. To this end, we first randomly sample 7 example images from the test set of the Stanford Image-Paragraph dataset~\cite{krause2017hierarchical}, and generate paragraphs using the baseline method called `Regions-Hierarchical'~\cite{krause2017hierarchical}. Additionally, we visualize the paragraph synthesized using our method under an ablation setting, \ie, where we don't have `Coherence Vectors' (indicated by `Ours (NC)'). This is followed by the synthesis results using our method trained under the regular non-VAE setting (indicated by `Ours'). Finally, we also present example visualizations for two paragraphs, per input image, generated by training our model under the VAE setting -- by choosing a different `z' each time (indicated by `Ours (with VAE) - I', and `Ours (with VAE) - II' respectively). 

Subsequently, we repeat a similar visualization exercise for the Amazon Product-Review Dataset~\cite{mcauley2015image} by choosing 15 random examples from the test set. However, while selecting these examples we ensure that we choose examples that span the different possible star ratings, \ie, from 1 star (indicating poor) all the way up to 5 stars (indicating good). Moreover, we also present paragraph synthesis results, for the scenario when the same product image is conditioned on different star ratings.

%\vspace{0.5cm}

\subsection{Stanford Image-Paragraph Dataset}
%\vspace{0.3cm}

\begin{figure}[h]
 \centering % 
 \includegraphics[height=16cm,width=\textwidth]{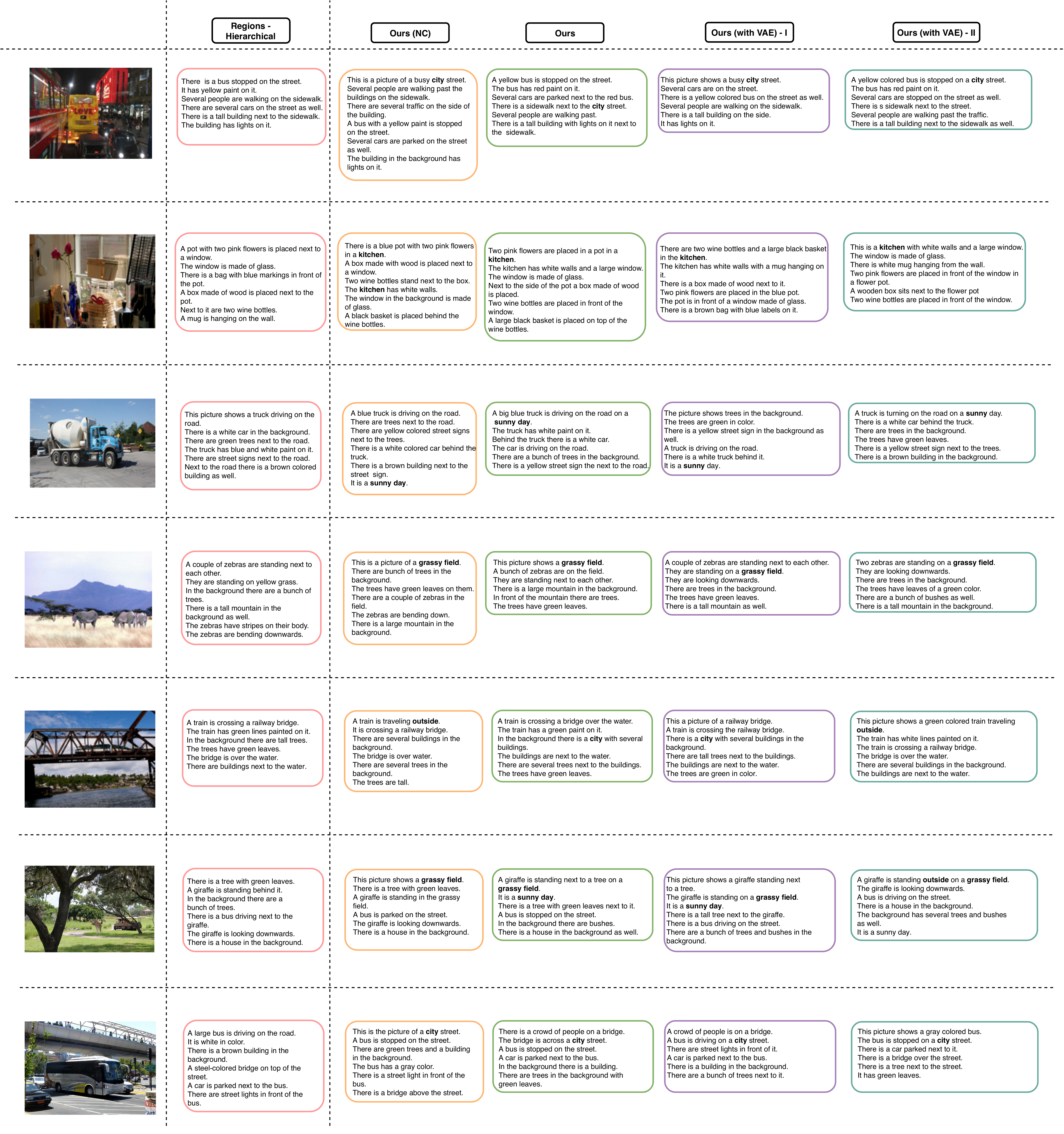}
 \caption{Paragraphs generated by using `Ours (NC),' `Ours,' `Ours (with VAE) - I,' `Ours (with VAE) - II' approaches, vis-\`a-vis the baseline method, Regions-Hierarchical~\cite{krause2017hierarchical}. The images are randomly sampled from the Stanford Image-Paragraph Dataset~\cite{krause2017hierarchical}. The words in bold are indicative of meta-concepts.} %competing methods.}
 \label{fig:samp_s1}
 \end{figure}

Results from the Stanford Image-Paragraph dataset~\cite{krause2017hierarchical} are shown in Figure~\ref{fig:samp_s1}. The text in bold in the figures indicates meta-concepts, like `city.' % ,~\ref{fig:samp_s2}
%{\color{blue}what's the difference between Fig.~1 and Fig.~2? Can we squeeze in more images? There is a lot of white space, \ie, wasted space, between the rows of the table in the figures.}
%\clearpage

%\begin{figure}[h]
% \centering % 
% \includegraphics[height=12cm,width=\textwidth]{Supplement_Stanford_Sample2.pdf}
% \caption{Paragraphs generated by using `Ours (NC),' `Ours,' `Ours (with VAE) - I,' `Ours (with VAE) - II' approaches, vis-\`a-vis the baseline method, Regions-Hierarchical~\cite{krause2017hierarchical}. The images are randomly sampled from the Stanford Image-Paragraph Dataset~\cite{krause2017hierarchical}. The words in bold are indicative of meta-concepts.} %competing methods.}
% \label{fig:samp_s2}
% \end{figure}

% {\color{blue}what do you mean by `under different settings'?}

%\vspace{0.5cm}

\subsection{Amazon Product-Review Dataset}
%\vspace{0.3cm}

\begin{figure}[h]
 \centering % 
 \includegraphics[height=16cm,width=\textwidth]{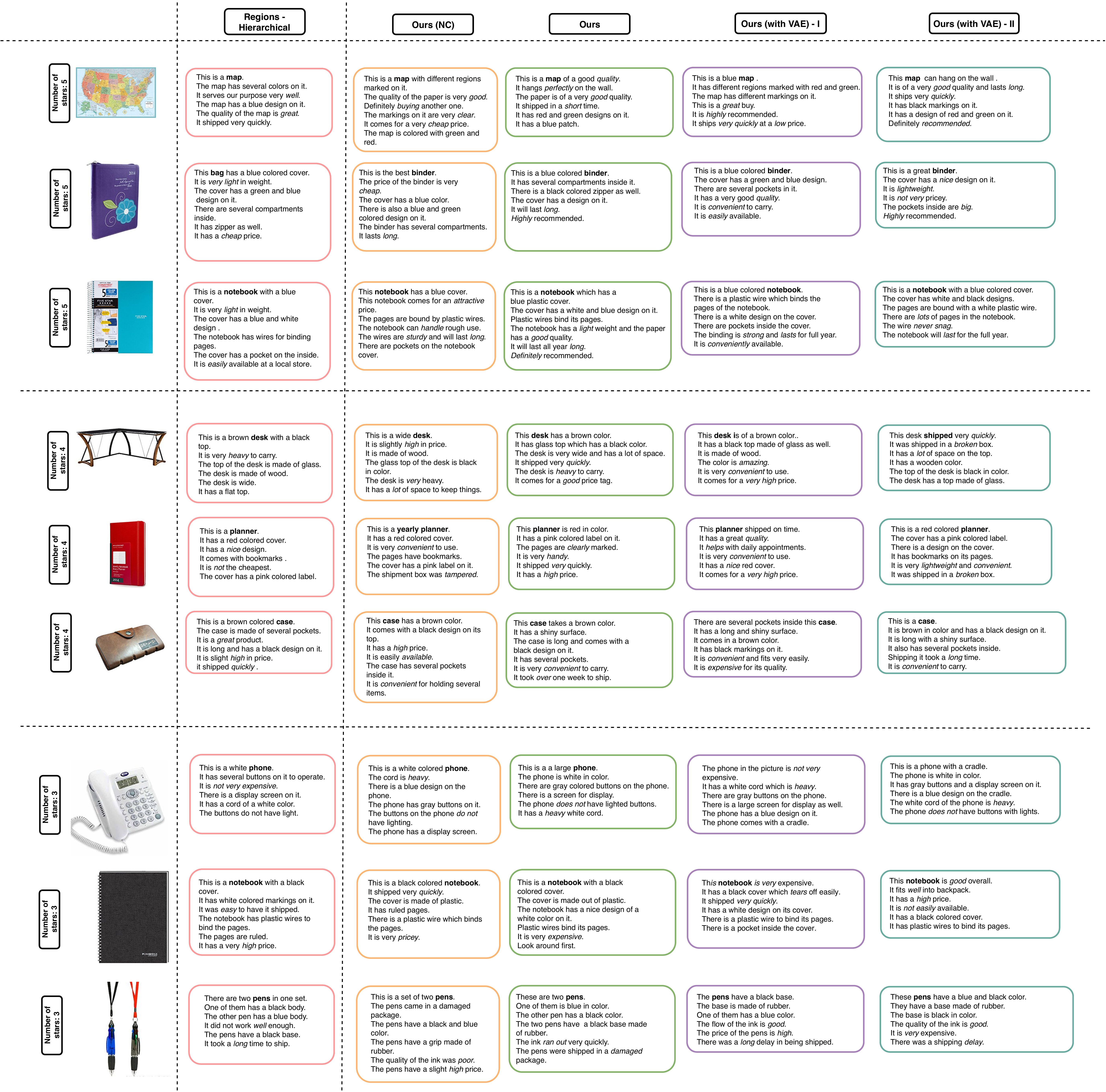}
 \caption{Paragraphs generated by using `Ours (NC),' `Ours,' `Ours (with VAE) - I,' `Ours (with VAE) - II' approaches, vis-\`a-vis the baseline method, Regions-Hierarchical~\cite{krause2017hierarchical}. The input images are conditioned on 3, 4, and 5 stars. The images are randomly sampled from the Amazon Product-Review Dataset~\cite{mcauley2015image}. The text in bold, indicates the principal object in the image, while the text in italics indicates words or phrases that are suggestive of different sentiments.} %competing methods.}
 \label{fig:samp_a1}
 \end{figure}

\begin{figure}[h]
 \centering % 
 \includegraphics[height=13cm,width=\textwidth]{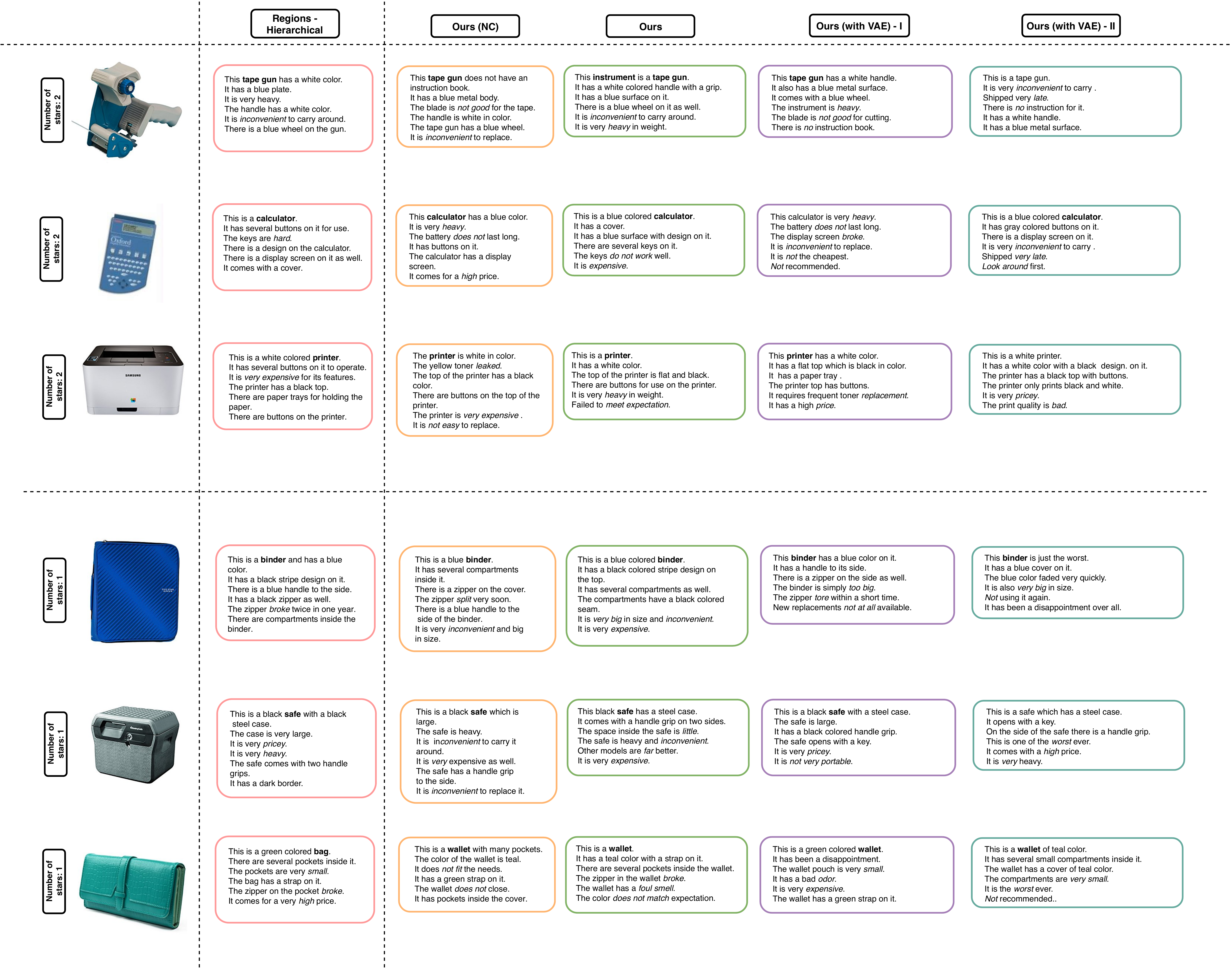}
 \caption{Paragraphs generated by using `Ours (NC),' `Ours,' `Ours (with VAE) - I,' `Ours (with VAE) - II' approaches, vis-\`a-vis the baseline method, Regions-Hierarchical~\cite{krause2017hierarchical}. The images are randomly sampled from the Amazon Product-Review Dataset~\cite{mcauley2015image}. The input images are conditioned on 1, and 2 stars. The text in bold, indicates the principal object in the image, while the text in italics indicates words or phrases that are suggestive of different sentiments.} %competing methods.}
 \label{fig:samp_a2}
 \end{figure}

Next, we present  results from the Amazon Product-Review dataset~\cite{mcauley2015image}. In Figure~\ref{fig:samp_a1}, we show results for input images conditioned with 3, 4, and 5 (good) star ratings, while in Figure~\ref{fig:samp_a2}, we present the results for the input images conditioned with 1 (poor), and 2 star ratings.  The text in bold in the figures indicates the principal object in the image, like `map,' while the text in italics indicates words or phrases that are suggestive of different sentiments, for instance, `inconvenient,' `great,' \etc %Both figures contain images for different star ratings, \ie 1 (indicating poor) through 5 (indicating good).
%{\color{blue}same comments as above apply.}
%\clearpage

\begin{figure}[h]
 \centering % 
 \includegraphics[height=12.5cm,width=\textwidth]{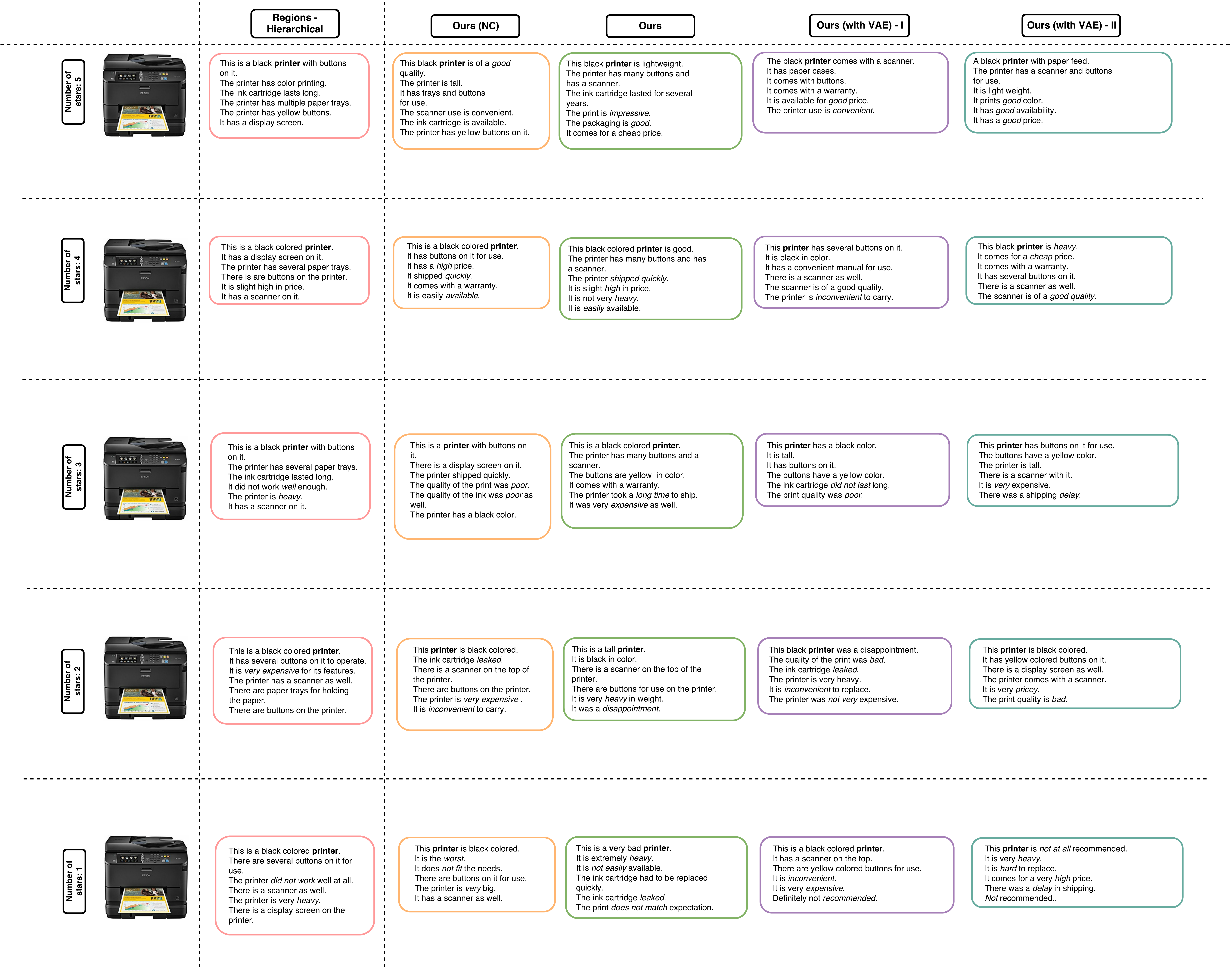}
 \caption{Paragraphs generated by using `Ours (NC),' `Ours,' `Ours (with VAE) - I,' `Ours (with VAE) - II' approaches, vis-\`a-vis the baseline method, Regions-Hierarchical~\cite{krause2017hierarchical}. The same product is conditioned on 1 through 5 star rating. The product is randomly sampled from the Amazon Product-Review Dataset~\cite{mcauley2015image}. The text in bold, indicates the principal object in the image, while the text in italicized font indicates words or phrases that are suggestive of different sentiments.} %competing methods.}
 \label{fig:samp_a3}
 \end{figure}

 We next compare the paragraph synthesis results for the same  product from the Amazon dataset~\cite{mcauley2015image} by our algorithm vis-\`a-vis `Regions-Hierarchical,'~\cite{krause2017hierarchical} when conditioned on different input star ratings. For purposes of this visualization, we show results for each of the possible star ratings, \ie, 1 (poor) through 5 (good). The results for this visualization are presented in Figure~\ref{fig:samp_a3}.
 
%\vspace{0.3cm}

\section{Derivation of the Coupling Unit Formulation}
%\vspace{0.1cm}
% Show the proof of the derivation here.

For completeness, in this section, we derive the closed form solution to the optimization problem that represents the \textit{Coupling Unit} of the \textit{Sentence Generation Net}.

The objective of the coupling unit is formulated as:
\[
%\underset{T^C_{i+1}}{\text{min}}
T^C_{i} = \arg\min_{\hat T^C_{i}}
 \quad \alpha \lvert\lvert T_{i} - \hat T^C_{i} \rvert\rvert^2_2 + \beta \lvert\lvert C_{i-1} - \hat T^C_{i} \rvert\rvert^2_2 \quad \text{with}\quad \alpha, \beta \geq 0,
\]
where both $\alpha, \beta$ are not simultaneously equal to $0$. 

The above objective is quadratic in nature and for the optimum $T^C_{i}$  a closed form solution is available. It is obtained by taking the gradient of the above objective with respect to $\hat T^C_{i}$ and setting the result to 0. In the following, we proceed with this strategy to derive the optimum $T^C_{i}$.

From
$$
f = \alpha \lvert\lvert T_{i} - \hat T^C_{i} \rvert\rvert^2_2 + \beta \lvert\lvert C_{i-1} - \hat T^C_{i} \rvert\rvert^2_2,
 $$
 we obtain after re-arranging terms %reformulation
 % \nabla_{\hat T^C_{i}}
%\text{or, }  \textit{f} = &\alpha (\lvert \lvert T_{i} \rvert \rvert^2 +  \lvert \lvert {\hat T^C_{i}} \rvert \rvert^2 - 2 <T_{i}, {\hat T^C_{i}}>) \quad \\
%+ &\beta (\lvert \lvert C_{i-1} \rvert \rvert^2  +  \lvert \lvert {\hat T^C_{i}} \rvert \rvert^2 - 2 <C_{i-1}, {\hat T^C_{i}}>) \\
$$
f = (\alpha + \beta) \lvert \lvert \hat T^C_{i} \rvert \rvert^2 - 2 (\alpha T_{i} + \beta C_{i-1})^T {\hat T^C_{i}} + \alpha \lvert \lvert T_{i} \rvert \rvert^2 + \beta \lvert \lvert C_{i-1} \rvert \rvert^2.
$$
Taking the gradient of $f$ with respect to $\hat T^C_{i}$ and setting the result to 0, \ie, solving
%
%\begin{align*}
%\text{So, } \nabla_{\hat T^C_{i}}\textit{f} = &0 \\
%\text{or, } 
$$
\nabla_{\hat T^C_{i}} f =  2 (\alpha + \beta) {\hat T^C_{i}} - 2 (\alpha T_{i} + \beta C_{i-1}) = 0
$$
\wrt $\hat T^C_i$ we obtain the solution 
%
%\text{or, } {\hat T^C_{i}} = & \frac{\alpha T_{i} + \beta C_{i-1}} {\alpha + \beta}
%\end{align*}
%
%Thus we have the optimum solution as :
\[%\begin{align*}
\boxed{T^C_{i} = \frac{\alpha T_{i} + \beta C_{i-1}} {\alpha + \beta}.}
\] %\end{align*}

\end{document}

%% file: introduction.tex
% !TEX root = main.tex
\section{Introduction}
Daily, we effortlessly describe fun events to friends and family, showing them pictures to underline the main plot. The narrative ensures that our audience can follow along step by step and picture the missing pieces in their mind with ease. Key to filling in the missing pieces is a consistency in our narrative which generally follows the arrow of time. 

%Daily, we effortlessly describe different scenarios to friends, family, and acquaintances, with the aid of pictures, and an accompanying narrative. The pictures typically underline the main plot, while the narrative underscores, in sufficient detail, the principal takeaways. Our audience can follow along this narrative, step by step, allowing their imaginative capacity to fill in the missing pieces with ease. Key to conveying this train of thought is a consistency in our narrative which generally follows the arrow of time. This raises a fundamental research question, \textit{``Can artificial intelligence (AI) techniques be used to automatically generate such coherent, detailed narratives, given an image?''}%Ensuring such a form of consistency in automatic paragraph generation has not yet been the main focus in our research community.

While computer vision, natural language processing and artificial intelligence techniques, more generally, have made great progress in describing visual content via image or video captioning~\cite{chen2015mind,donahue2015long,karpathy2015deep,mao2014deep,vinyals2015show}, the obtained result is often a single sentence of around 20 words, describing the main observation. Even if brevity caters to today's short attention span, 20 words are hardly enough to describe subtle interactions, let alone detailed plots of our experience. Those are much more meaningfully depicted in a paragraph of reasonable length.

%This research question is of keen interest to the AI community, since it requires sophisticated techniques that marry the two fields of computer vision, and natural language processing. With the advent of large datasets that couple images with ground-truth human annotation~\cite{hodosh2013framing,krishna2017visual,lin2014microsoft,young2014image}, several new algorithms have been proposed to describe the visual content via automated captioning~\cite{chen2015mind,donahue2015long,karpathy2015deep,mao2014deep,vinyals2015show}. These algorithms typically generate captions that describe the visual content, at a coarse-level of detail, in a single sentence of around 20 words. Even if brevity caters to today's prevalent short attention span, a single sentence of 20 words is hardly enough to describe subtle interactions between the entities appearing in the image, let alone tie them in one coherent plot. This necessitates the generation of paragraphs of reasonable length, to describe the image.
 
%To this end, visual paragraph generation methods~\cite{} aim to provide a longer narrative which describes a given image or video. Early existing methods~\cite{} attempted to create a single paragraph for a given input. While classical machine learning techniques are applicable in this case, the ambiguity of the task and hence the expressivity is lost when using models of this type. Therefore, more recently, diverse paragraph generation has been considered by {\color{blue}xxx}~\cite{}. The approaches {\color{blue}description here}.

To this end, visual paragraph generation methods~\cite{johnson2016densecap,krause2017hierarchical,yu2016video,liang2017recurrent}, which have been proposed very recently,  provide a longer narrative which describes a given image or video. However, as argued initially, coherence between successive sentences of the narrative is a key necessity to effectively convey the plot of our experience. Importantly, models for many of the aforementioned methods provide no explicit mechanisms to ensure cross-sentence topic consistency. A notable exception is the work of Liang \etal.~\cite{liang2017recurrent}. 

In particular, Liang \etal.~\cite{liang2017recurrent} propose to ensure consistency across sentence themes by training a standard paragraph generation module~\cite{krause2017hierarchical}, coupled with an attention mechanism, under a Generative Adversarial Network (GAN)~\cite{goodfellow2014generative} setting which has an additional loss-term to enforce  consistency. However, difficulties associated with training GANs~\cite{arjovsky2017wasserstein} and no explicit coherence model, leave their method vulnerable to generating incoherent paragraphs.
%This is especially true, in this case, since GANs are known to be hard to train~\cite{arjovsky2017wasserstein}. 
%{\color{blue}what is bad about this that we fix?} %However,~\cite{krause2017hierarchical,johnson2016densecap}  are capable of generating only a single paragraph for a given input. {\color{blue}we need to show multiple paragraphs if we say this.} %This compromises on the expressivity of the model. Therefore, 

%More recently, diverse paragraph generation has gained traction~\cite{liang2017recurrent}. Their approach casts the synthesis task, in a Generative Adversarial training setting, using {\color{blue}xxx} to ensure consistency between sentences. However, {\color{blue}what is bad about this that we fix?} %, which is known to be challenging~\cite{arjovsky2017wasserstein}. %Differently, our algorithm ensures diversity of the synthesized paragraphs by incorporating our sentence generation scheme in a Variational Autoencoder framework~\cite{kingma2013auto} that suffers from none of the aforementioned issues.

 \begin{figure}[t]
 \centering
 \includegraphics[width=\textwidth]{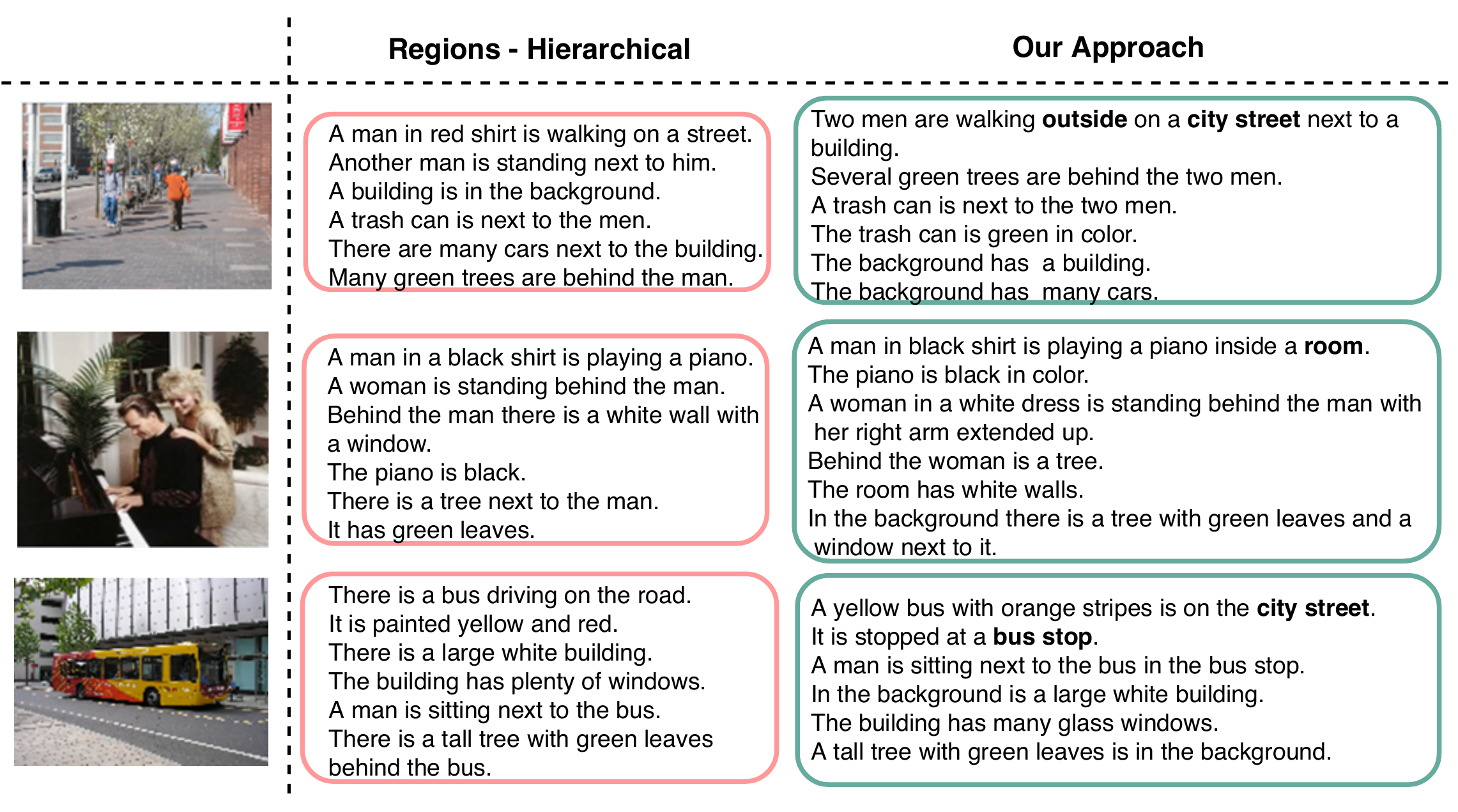}
 \caption{Paragraphs generated with a prior state-of-the-art technique~\cite{krause2017hierarchical} and with our developed approach. Due to the introduced `Coherence Vectors'  we observe the generated paragraphs to be much more coherent than prior work~\cite{krause2017hierarchical}}. %{\color{blue}too much white-space between the bottom of the figure and the caption; the dashed line should end right at the bottom of the last row's box; why waste space; add one more row of examples; change `Our proposed approach' to `Our Approach'}}

 %} %competing methods.}
 \label{fig:teaser}
 \end{figure}
 
 %As argued initially, consistency of the narrative is a necessity for painting pictures in the mind of the audience. Importantly, many of the aforementioned methods provide no explicit mechanisms in the developed models to ensure consistency. Notable exceptions are~\cite{}, which {\color{blue}xxx}.
 
% In contrast, in this work, we explicitly focus on modeling of consistency for paragraph generation and develop a model that propagates what we call `consistency vectors.' More specifically, our developed model {\color{blue}does xxx}. We illustrate provided images and generated paragraphs in \figref{fig:teaser} showing {\color{blue}xxx}.

%In contrast, we address the challenge of coherent visual paragraph generation by hierarchically synthesizing topic vectors for every sentence. %, followed by the synthesis of the actual sentence~\cite{krause2017hierarchical}. 
%Different from all prior work, our model benefits from coupling these synthesized sentence-level topic vectors, with a ``\textit{Global Topic Vector},'' which encapsulates a summary of the image content. Further a key component of our model, in contrast to 
Different from  prior work, we explicitly focus on modeling the \emph{diverse yet coherent possibilities}  of successive sentences when generating a paragraph, while ensuring that the `big picture' underlying the image does not get lost in the details. To this end, we develop a model that propagates, what we call ``\textit{Coherence Vectors},'' which ensure cross-sentence topic smoothness, and a  ``\textit{Global Topic Vector},'' which captures the summarizing information about the image. Additionally, we observe improvements in the quality of the generated paragraphs, when our model is trained to incorporate diversity. %, while generating paragraphs.
%Beyond ensuring  coherence the are encoding a \emph{diverse} set of possibilities to continue a sentence. 
Intuitively, the coherence vector embeds the theme of the most recently generated sentence. 
{%The topic vector of the next sentence is combined with the coherence vector from the most recently generated one and the global topic vector to generate a new topic vector, with the intention to ensure a smooth flow of the theme across sentences. 
To ensure a smooth flow of the theme across sentences, we combine the coherence vector with the topic vector of the current sentence and a global topic vector. % and a coherence vector. %The latter captures data about the previously generated sentence. 
Figure~\ref{fig:teaser} illustrates a sampling of a synthesized paragraph given an input image, using our method vis-\`a-vis prior work~\cite{krause2017hierarchical}. Notably, using our model, we observe a smooth transition between sentence themes, while capturing summarizing information about the image. For instance, generated paragraphs corresponding to the images in the first and the third rows in Figure~\ref{fig:teaser} indicate that the images have been captured in a `city' setting.} 
%Additionally, we observe synergistic improvements in the quality of the gnerated paragraphs, when our model is trained to incorporate diversity.
%{\color{red}I'll fix the preceding sentence once seeing the figure}

%Moreover, existing methods~\cite{krause2017hierarchical,johnson2016densecap}, lack variety, and are capable of generating only a single paragraph for a given input. This compromises on the expressivity of the model. Therefore, more recently, diverse paragraph generation has gained traction~\cite{liang2017recurrent}. Their approach casts the synthesis task, in a Generative Adversarial training setting, which is known to be prone to training difficulties~\cite{arjovsky2017wasserstein} and also doesn't support straight-forward back-propagation based training due to the presence of an intermediate sampling step. Differently, our algorithm ensures diversity of the synthesized paragraphs by incorporating our sentence generation scheme in a Variational Autoencoder framework~\cite{kingma2013auto} that suffers from none of the aforementioned issues.

%Following existing work we evaluate the developed approach on {\color{blue}xxx}~\cite{}, demonstrating state-of-the-art performance which outperforms existing methods by {\color{blue}xxx}\%. Different from existing methods we also evaluate the proposed approach on {\color{blue}xxx} showing {\color{blue}xxx}. We think    evaluation on this additional dataset provides important mechanisms to safe-guard against overfitting.

Following prior work we quantitatively evaluate our approach on the standard Stanford Image-Paragraph dataset~\cite{krause2017hierarchical}, demonstrating state-of-the-art performance. Furthermore, different from all existing methods, we showcase the generalizability of our model, evaluating the proposed approach by generating reviews from the ``Office-Product'' category of the Amazon product review dataset~\cite{mcauley2015image} and by showing significant gains over all baselines. 

In the next section, we discuss prior relevant work before providing details of our proposed approach in \secref{sec:app}. \secref{sec:exp} presents empirical results. We finally conclude in \secref{sec:conc}, laying out avenues for future work.

%% file: related.tex
% !TEX root = main.tex
\section{Related Work}
% Comments by Alex - 

%Related to our developed approach of paragraph generation with `consistency vectors' is work on image captioning, paragraph generation, and language and vision tasks like visual question answering, visual dialog, and visual grounding in general. We review the latter more broadly and the former two in greater detail in the following.

%\paragraph{Language and Vision:} At the intersection of computer vision, natural language processing and machine learning {\color{blue}describe a few of the aforementioned tasks at the high level}

%Two important tasks that spurred much of the language and vision research today are image captioning and paragraph generation.

%\paragraph{Image Captioning:} {\color{blue}xxx}

%Our developed approach differs from the aforementioned image captioning techniques in that we generate a  long paragraph rather than a short caption. From a technical perspective, long-range consistency is not important when generating a single sentence.

%\paragraph{Visual Paragraph Generation:}  {\color{blue}xxx}

%Our devised method differs from the aforementioned visual paragraph generation techniques in that we introduce `consistency vectors,' which explicitly encourage long-range consistency between sentences of a paragraph. We introduce our `consistency vectors' in the following.

% Updated Related Works

For a long time, associating language with visual content has been an important research topic~\cite{lavrenko2004model,xiao2007fusion,chatterjee2015novel}.  Early techniques in this area associate linguistic `tag-words' with visual data. %We, on the other hand, concern ourselves with associating full-blown sentences/paragraphs with visual data. 
Gradually, the focus shifted to generating entire sentences and paragraphs for visual data. For this, techniques from both natural language processing and computer vision are combined with the aim of building holistic AI systems that integrate naturally into common surroundings. Two tasks that spurred the growth of recent work in the language-vision area are \textit{Image Captioning}~\cite{vinyals2015show,johnson2016densecap,chen2015mind,donahue2015long,xu2015show}, and \textit{Visual Question Answering}~\cite{antol2015vqa,gao2015you,shih2016look,ren2015exploring,malinowski2015ask,xiong2016dynamic,xu2016ask,yang2016stacked,SchwartzNIPS2017,JainCVPR2018}. More recently, image captioning approaches were extended to generate natural language descriptions at the level of paragraphs~\cite{krause2017hierarchical,johnson2016densecap,liang2017recurrent}. In the following, we review related work from the area of image captioning and visual paragraph generation in greater detail, and point out the distinction with our work. %Prior to that, we provide a broader survey of language and vision tasks.

\paragraph{\textbf{Image Captioning:}} \textit{Image Captioning} is the task of generating textual descriptions, given an input image. Classical methods for image captioning, are usually non-parametric.  These methods build a pool of candidate captions from the training set of image-caption pairs, and at test time, a fitness function is used to retrieve the most compelling caption for a given input image~\cite{lavrenko2004model,pan2004automatic,chatterjee2015novel}. However the computationally demanding nature of the matching process imposes a  bottleneck when considering a set of descriptions of a reasonable size. 

To address this problem, Recurrent Neural Network (RNN)-based approaches have come into vogue~\cite{vinyals2015show,mao2014deep,you2016image,xu2015show,karpathy2015deep,WangNIPS2017,AnejaCVPR2018,DeshpandeARXIV2018} lately. These approaches, typically, first use a Convolutional Neural Network (CNN)~\cite{simonyan2014very,krizhevsky2012imagenet} to obtain an encoding of the given input image. This encoding is then fed to an RNN which samples a set of words (from a dictionary of words) that agree  most with the image encoding. However, the captions generated through such techniques are short, spanning typically a single sentence of at most 20 words. Our approach differs from the aforementioned image captioning techniques, in that we generate a  paragraph of multiple sentences  rather than a short caption. Importantly, captioning techniques generally don't have to consider coherence across sentence themes, which is not true for paragraph generation approaches which we review next.
% This generation is  performed in a sequential manner, word-by-word.

\paragraph{\textbf{Visual Paragraph Generation:}}  From a distance, the task of \textit{Visual Paragraph Generation} resembles image captioning: given an image, generate a textual description of its content~\cite{krause2017hierarchical}. However, of importance for visual paragraph generation is the attention to detail in the textual description. In particular, the system is expected to generate a paragraph of sentences (typically 5 or 6 sentences per paragraph) describing the image in great detail. Moreover, in order for the paragraph to resemble natural language, there has to be a smooth transition across the themes of the sentences of the paragraph.

Early work in generating detailed captions, include an approach by Johnson \etal\cite{johnson2016densecap}. While generating compelling sentences individually, a focus on a theme of the story underlying a given image was missing. This problem was addressed by Krause \etal~\cite{krause2017hierarchical}. Their language model consists of a two-stage hierarchy of RNNs. The first RNN level  generates sentence topics, given the visual representation of semantically salient regions in the image. The second RNN level translates this topic vector into a sentence. This model was further extended by Liang \etal~\cite{liang2017recurrent} to encourage coherence amongst successive sentences. To this end, the language generation mechanism of Krause \etal~\cite{krause2017hierarchical}, coupled with an attention mechanism, was trained in a Generative Adversarial Network (GAN) setting, where the discriminator is intended to encourage this coherence at training time. Dai~\etal~\cite{dai2017towards} also train a GAN for generating paragraphs. However, known difficulties of training GANs~\cite{arjovsky2017wasserstein} pose challenges towards effectively implementing such systems. Xie~\etal introduce regularization terms for ensuring diversity~\cite{xie2018thesis} which results in a constrained optimization problem that does not admit a closed form solution and is thus hard to implement. Different from these approaches~\cite{liang2017recurrent,dai2017towards,xie2018thesis}, we demonstrate that a change of the generation mechanism is better suited to obtain coherent sentence structure within the paragraph. To this end we introduce \textit{Coherence Vectors} which ensure a gradual transition of themes between sentences. 
%we fundamentally change the underlying paragraph generation scheme by permitting the flow of consistency vectors, to ensure coherence between the topics of successive sentences. 

Additionally, different from  prior work, we also incorporate a summary of the topic vectors to sensitize the model to the `main plot' underlying the image. Furthermore, to capture the inherent ambiguity of generating paragraph from images, \ie, multiple paragraphs can successfully describe an image, we cast our paragraph-generation model as a Variational Autoencoder (VAE)~\cite{kingma2013auto,jain2017creativity,chung2015recurrent,gregor2015draw}, enabling our model to generate a set of diverse paragraphs, given an image. 

%Our devised method differs from the aforementioned visual paragraph generation techniques in that we introduce `consistency vectors,' which explicitly encourage long-range consistency between sentences of a paragraph. We introduce our `consistency vectors' in the following.

%% file: method.tex
% !TEX root = main.tex
\section{Our Proposed Method for Paragraph Generation} 
%\section{A Method for Diverse and Coherent Paragraph Generation} %{Consistency Vectors for Paragraph Generation}
\label{sec:app}

\begin{figure}[t]
\centering
\includegraphics[scale = 0.5]{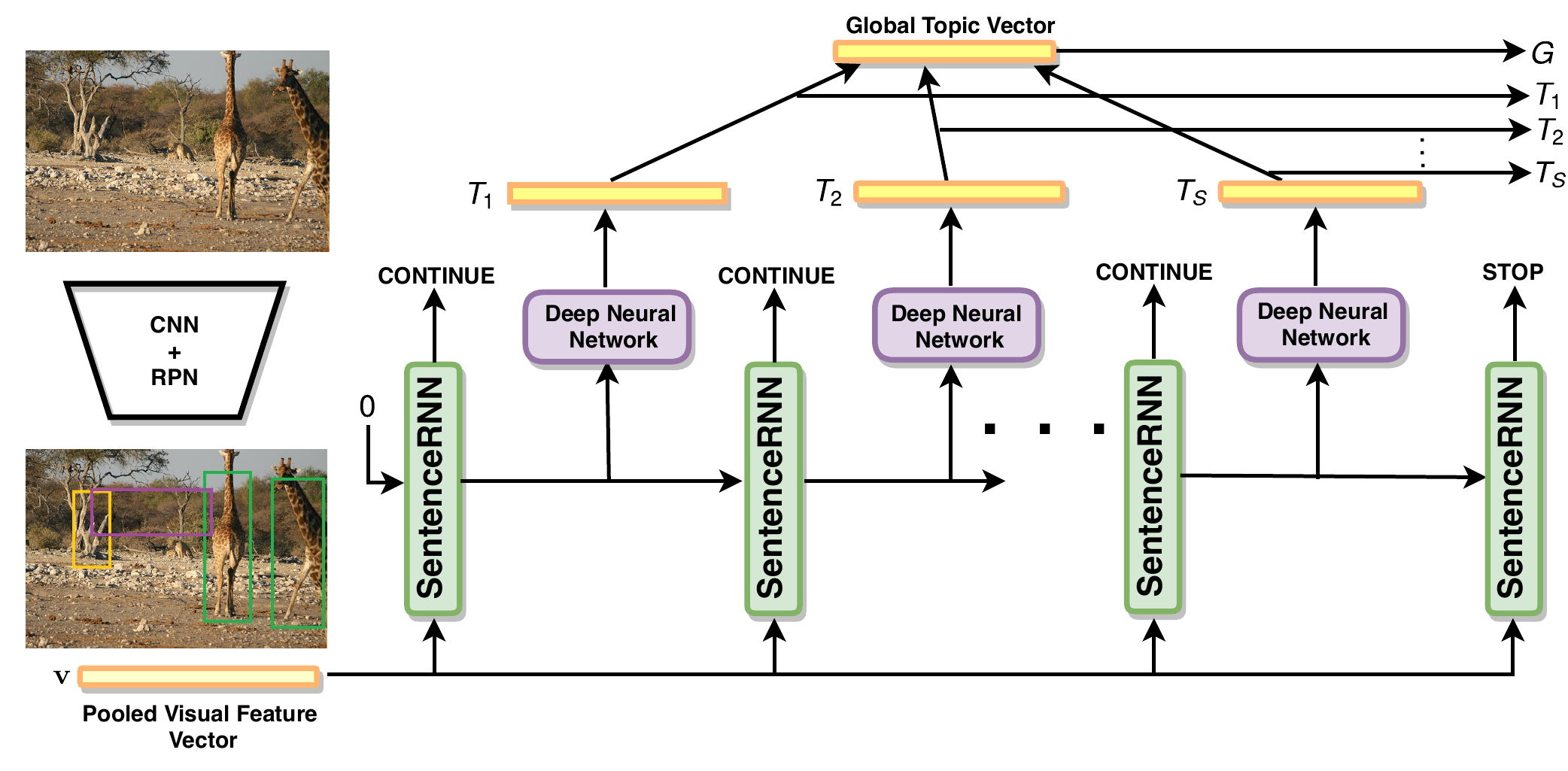}
\caption{Overview of the Topic Generation Net of our proposed approach illustrating the construction of the individual and `Global Topic Vector'.} %{\color{blue}please make all font sizes identical; this makes the figure look well organized; I think we talked about this - done. Two sets of font sizes were used, one for the SentenceRNN, and the other for the notations. Making all text have the same font size was resulting in a lot of white space in the senten; also move the purple `deep neural net' box a little to the right such that it's not as close to `CONTINUE' - done}}

\label{fig:topic_gen}
%\end{figure}
%\bigbreak

%\begin{figure} [h]
\centering
\includegraphics[scale = 0.7]{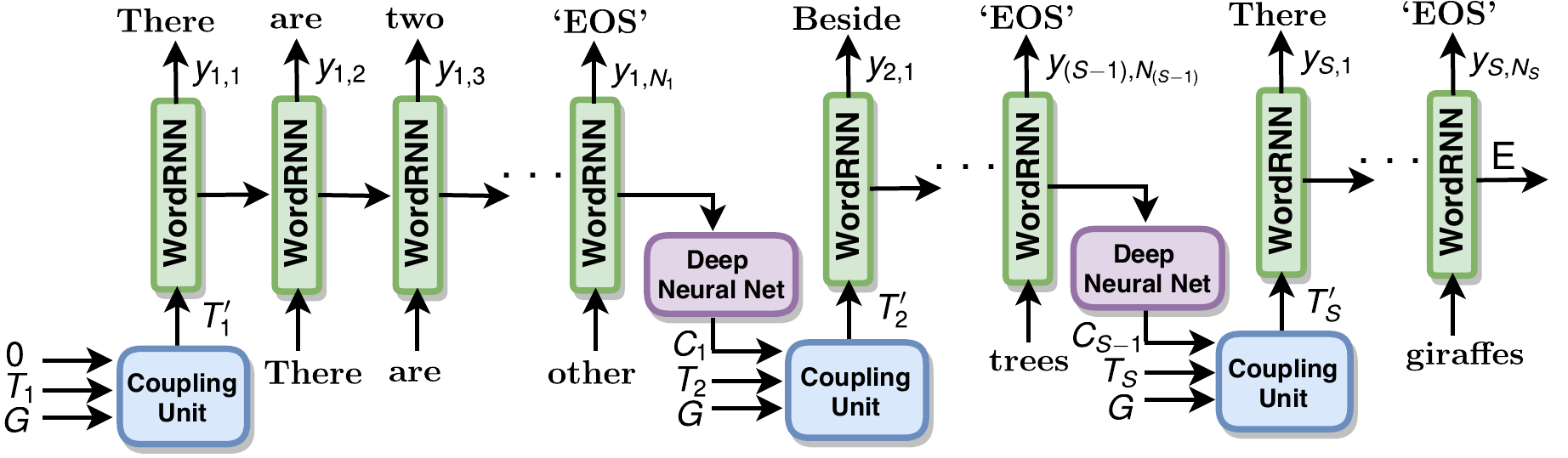}
\caption{Overview of the Sentence Generation Net.} %{\color{blue}make sure that the font sizes are all identical - fixed. I have used two sets of font sizes, one for mathematical notations and the other labeling the boxes. Using one font size was making it look weird; can we indicate what is $y_1$, $y_2$, etc.; or maybe $y_{ij}$ - done}} 

\label{fig:para_gen}
\end{figure}
%Of crucial importance for visual paragraph generation is consistency between the generated sentences. To explicitly encourage this consistency we introduce `consistency vectors.' In the following we provide an overview of the proposed approach before discussing the details more carefully.

As mentioned before, coherence of sampled sentences is important for automatic generation of human-like paragraphs from visual data, while not losing sight of the underlying `big picture' story illustrated in the image. Further, another valuable element for an automated paragraph generation system is the diversity of the generated text. In the following we develop a framework which takes into account these properties. We first provide an overview of the approach in \secref{sec:Overview}, before discussing our approach to generate coherent paragraphs in \secref{sec:Coherence} and finally our technique to obtain diverse paragraphs in \secref{sec:Diverse}.

%As mentioned before, consistency of sampled sentences is  important for automatic generation of human-like paragraphs from visual data, so as to not loose sight of the underlying `big picture' of the image. Further, another valuable element for an automated paragraph generation system is the diversity of the generated text. In the following we develop a framework which takes into account both properties. We first provide an overview of the proposed approach for incorporating these desirable elements into our system, before discussing the method in greater details.

\subsection{Overview}  
\label{sec:Overview}
% Alex - In Figure \ref{fig:overview} we illustrate an overview demonstrating the use of the introduced `consistency vectors.' {\color{blue}figure which illustrates the method and a corresponding description}

%{\color{red}we should introduce the output space symbols, \ie, words here and provide the training loss in Section 3.2. such that it's consistent with what follows in Section 3.3; right now the story is not coherent}

%Paragraph generation datasets can be mathematically represented as a set of $(x, y)$ pairs, where $x$ is the image, and $y$ is the ground-truth paragraph. A paragraph $y$ consists of $S$ sentences, with the $i^{th}$ sentence having $N_i$ words, and $y_{ij}$ denoting the the $j^{th}$ word of the $i^{th}$ sentence. We leverage such training data to train our system, an overview of whose architecture follows next.

To generate a paragraph $y = (y_1, \ldots, y_S)$ consisting of $S$ sentences $y_i$, $i\in\{1, \ldots, S\}$, each with $N_i$ words $y_{i,j}$, $j\in\{1, \ldots, N_i\}$, for an image $x$, we  use  a deep net composed out of two  modules which are coupled hierarchically: the \textit{Topic Generation Net} and the \textit{Sentence Generation Net}. %Both deep nets are trained end-to-end using labeled training data, which consists of pairs $(x,y)$ of images $x$ and a corresponding paragraph $y$. %We first provide an overview of both deep nets, and the training technique subsequently.

%Each paragraph consists of $S$ sentences, $y = (y_1, \ldots, y_S)$, where each sentence $y_i$ consists of $N_i$ words, $y_i = (y_{i1}, \ldots, y_{iN_i})$

%Figure~\ref{fig:topic_gen} presents an overview of
The \textit{Topic Generation Net} illustrated in Figure~\ref{fig:topic_gen} seeks to extract a set of $S$ topic vectors, $T_i \in \mathbb{R}^H$ $\forall i\in\{1, \ldots, S\}$, given an appropriate visual representation of the input image $x$. The topic generation net is a parametric function which, recursively at every timestep, produces a topic vector $T_i$ and a probability measure $u_i$ indicating if more topics are to be generated. We implement this function using a recurrent net, subsequently also referred to as the \textit{SentenceRNN}.
We then leverage the topic vectors $T_i$ to construct a \textit{Global Topic Vector} $G \in \mathbb{R}^H$, which captures the underlying image summary. This global topic vector is constructed via a weighted combination of the aforementioned topic vectors $T_i$.

Figure~\ref{fig:topic_gen} illustrates a detailed schematic of the topic generation net. %We discuss the details of the generation scheme subsequently.
Formally we use $(G, \{(T_i, u_i)\}_{i=1}^S) = \Gamma_{w_T}(x)$ to denote the input and output of the net $\Gamma_{w_T}(\cdot)$, where the vector $w_T$ subsumes the parameters of the function. 
The global topic vector $G$, and the individual topic vectors and probabilities $\{(T_i, u_i)\}_{i=1}^S$ are the output %of this \textit{Topic Generation Net} and 
which also constitute the input to the second module.

%\paragraph{\textbf{Overview:}} 
%Figure~\ref{fig:topic_gen} presents an overview of the \textit{Topic Generation Net}, which seeks to extract a set of $S$ topic vectors corresponding to different semantically salient regions detected in the given input image $x$.  To extract the topic vectors we use a recurrent net subsequently referred to as the \textit{SentenceRNN}, which produces the topic vectors $T_i \in \mathbb{R}^H$ $\forall i\in\{1, \ldots, S\}$. We then leverage these topic vectors to construct a \textit{Global Topic Vector} $G \in \mathbb{R}^H$, which captures the underlying image summary. This global vector is constructed via a weighted combination of the aforementioned topic vectors $T_i$. We discuss the details of the generation scheme subsequently. The global topic vector $G$, and the individual topic vectors ($\{T_i\}_{i=1}^S$) are the output of this \textit{Topic Generation Net} and  the input of the second module.

The second module of the developed approach, called the \textit{Sentence Generation Net}, is illustrated in Figure~\ref{fig:para_gen}. Based on the output of the topic generation net, it is responsible for producing a paragraph $y$, one sentence $y_i$ at a time. %A sentence, $y_i$, $i\in\{1, \ldots, S\}$, is generated based on every topic vector $T_i$.

Formally, the sentence generation module is also modeled as a parametric function which synthesizes a sentence $y_i$, one word $y_{i,j}$ at a time. More specifically, a recurrent net $\Gamma_{w_\text{s}}(\cdot, \cdot)$ is used to obtain the predicted word probabilities $\{p_{i,j}\}_{j=1}^{N_i} = \Gamma_{w_\text{s}}(T_i, G)$, where $w_\text{s}$ subsumes all the parameters of the net, and $p_{i,j} \in [0,1]^V$ $\forall j\in\{1, \ldots, N_i\}$ is a probability distribution over the  set of $V$ words in our vocabulary (including an `End of Sentence' (`EOS') token). We realize the function, $\Gamma_{w_\text{s}}(\cdot,\cdot)$ using a recurrent net, subsequently referred to as the \textit{WordRNN}.

In order to incorporate cross-sentence coherence, rather than directly using the topic vector $T_i$ in the WordRNN, we first construct a modified topic vector $T'_i$, which better captures the theme of the $i^{th}$ sentence. For every sentence $i$, we compute $T'_i \in \mathbb{R}^H$ via a \textit{Coupling Unit}, by combining the topic vector $T_i$, the global topic vector $G$ and a previous sentence representation $C_{i-1}$, called a \textit{Coherence Vector}, which captures properties of the sentence generated at step $i-1$. Note that the synthesis of the first sentence begins by constructing $T'_1$, which is obtained by coupling $T_1$ with the global topic vector $G$, and an all zero vector. %The collection of parameters of all other parameterized components of our system, besides the WordRNN or SentenceRNN, such as the coupling unit are collectively denoted by $w_X$. % to denote , other than the wordRNN or sentenceRNN, such as the coupling unit. %{\color{blue}what do we need $w_X$ for? this is strange, particularly if you don't say which parameters are in there?}  % and the details follow subsequently. We begin by discussing the technique for extracting a representation of the image $x$.
\paragraph{\textbf{Visual Representation:}} 
To obtain an effective encoding of the input image, $x$, we follow Johnson \etal~\cite{johnson2016densecap}. More specifically, a Convolutional Neural Network (CNN) (VGG-16~\cite{simonyan2014very}) coupled with a Region Proposal Network (RPN) gives fixed-length feature vectors for every detection of a semantically salient region in the image. The obtained set of vectors $\{v_1, \ldots, v_M\}$ with $v_i \in \mathbb{R}^D$ each correspond to a region in the image. We subsequently pool these vectors into a single vector, $v \in \mathbb{R}^I$ -- following the approach of Krause \etal~\cite{krause2017hierarchical}. This pooled representation contains relevant information from the different semantically salient regions in the image, which is supplied as input to our topic generation net. Subsequently, we use $v$ and $x$ interchangeably. 
% , by coupling a fully connected layer, along with a maxpool layer 
\subsection{Coherent Paragraph Generation}
\label{sec:Coherence}
The construction of coherent paragraphs adopts a two-step approach. In the first step, we derive a set of individual and a global topic-vector starting with the pooled representation of the image. This is followed by paragraph synthesis. % of the paragraph. %We then adopt a hierarchical RNN-based approach towards synthesizing the sentences of the paragraph in a coherent fashion, which we discuss next. %. In the following we  discuss the details of this framework.
%that first generates topic vectors, combines them with representations of previously generated sentences into modified topic vectors which are subsequently translated into sentences. 

\paragraph{\textbf{Topic Generation:}}
%\begin{figure}[t]
%\centering
%\includegraphics[height=3.5cm,width=10cm]{fig/First_Pass_Glob.png}
%\caption{Mechanism for constructing global vectors.}
%\label{fig:glob}
%\end{figure}
The \textit{Topic Generation Net} $(G, \{(T_i, u_i)\}_{i=1}^S) = \Gamma_{w_T}(x)$ constructs a set of relevant topics $T_i$ for subsequent paragraph generation given an image $x$. Figure~\ref{fig:topic_gen} provides a schematic of the proposed topic generation module. At first, the pooled visual representation of the image, $v$, is used as input for the \textit{SentenceRNN}. %{\color{blue} you are referring to this RNN with various notations sentenceRNN, SentenceRNN, Sentence RNN; make it consistent; same for wordRNN} %~\cite{krause2017hierarchical,liang2017recurrent}. 
The SentenceRNN is a single layer Gated Recurrent Unit (GRU)~\cite{chung2014empirical}, parameterized by $w_T$. It takes an image representation $v$ as input and produces a probability distribution $u_i$, over the labels `CONTINUE' or `STOP,' while its hidden state is used to produce the topic vector $T_i \in \mathbb{R}^H$ via a 2-layer densely connected deep neural network. A `CONTINUE' label ($u_i > 0.5$), indicates that the recurrence should proceed for another time step, while a `STOP' symbol terminates the recurrence. 
%The hidden state of the GRU, at step $i$, is used to produce a topic vector $T_i \in \mathbb{R}^H$, via a 2-layer densely connected deep neural network. This topic vector is eventually used to generate the $i^{th}$ sentence. This process continues, until a `STOP' is predicted.

However, automatic description of an image via paragraphs necessitates tying all the sentences of the paragraph to a `big picture' underlying the scene. For example, in the first image in Figure~\ref{fig:teaser}, the generated paragraph should ideally reflect that it is an image captured in a `city' setting. To encourage this ability we construct a \textit{Global Topic Vector} $G \in \mathbb{R}^H$ for a given input image (see Figure~\ref{fig:topic_gen}). %This constitutes a principal novelty of our approach.

Intuitively, we want this global topic vector to encode a holistic understanding of the image, by combining the aforementioned individual topic vectors as follows:
\be
G = \sum_{i = 1}^n \alpha_i T_i \quad \text{where} \quad \alpha_i = \frac{\lvert\lvert T_i \rvert\rvert_2}{\sum_i \lvert\lvert T_i \rvert\rvert_2}.
\ee
Our intention is to facilitate representation of `meta-concepts' (like `city') as a weighted combination of its potential constituents (like `car,' `street,' `men,' \etc). The synthesized global vector and the topic vectors are then propagated to the sentence generation net which predicts the words of the paragraph. % generating the paragraph.

\paragraph{\textbf{Sentence Generation:}}
Given the individual topic vectors $T_i$ and the global topic vector $G$, the \textit{Sentence Generation Net} synthesizes sentences of the paragraph by computing word probabilities $\{p_{i,j}\}_{j=1}^{N_i} = \Gamma_{w_\text{s}}(T_i, G)$, conditioned on the previous set of synthesized words %. To this end, for generation of a sentence $y_i$  a representation of the previous sentence $C_{i-1}$, the topic vector $T_i$ and the global topic vector $G$ via the coupling unit 
(see Figure~\ref{fig:para_gen}). 
One sentence is generated for each of the $S$ individual topic vectors $T_1, \ldots,T_S$. Synthesis of the $i^{\text{th}}$ sentence commences by  combining via the \textit{Coupling Unit} the topic vector $T_i$,  the global topic vector $G$, and the consistency ensuring \textit{Coherence Vector} $C_{i-1} \in \mathbb{R}^H$. %The \textit{Coherence Vector} $C_{i-1}$ is obtained by transforming the last hidden state of the WordRNN from the $(i-1)^{th}$ sentence via a deep net. 

The \textit{Coupling Unit} produces a modified topic vector ($T^\prime_i \in \mathbb{R}^H$), which is propagated to the \textit{WordRNN} to synthesize the sentence. The \textit{WordRNN} is a 2-layer GRU, which generates a sentence, $y_i$, one word at a time, conditioned on the previously synthesized words. The $j^{th}$ word of the $i^{th}$ sentence is obtained by selecting the word with the highest posterior probability, $p_{i,j}$, over the entries of the vocabulary $V$. A sentence is terminated when either the maximum word limit per sentence is reached or an `EOS' token is predicted. In the following, we describe the mechanism for constructing the coherence vectors, and the coupling technique referenced above.

\paragraph{Coherence Vectors:}
An important element of human-like paragraphs is coherence between the themes of successive sentences, which ensures a smooth flow of the line of thought in a paragraph.
%Most prior work for paragraph generation from images does not explicitly encourage this cross-sentence coherence~\cite{johnson2016densecap,krause2017hierarchical}. A notable exception is the work by Liang \etal~\cite{liang2017recurrent}. To encourage topic-coherence, the authors introduce a topic-transition loss to the discriminator component of their generative adversarial net (GAN), while essentially using the basic paragraph generation module of Krause \etal~\cite{krause2017hierarchical}, coupled with an attention mechanism. However, topic coherence is not built into the generator architecture, leaving the generator module prone to incoherent paragraph generation, especially if the generator is improperly trained. 

%%Our approach diverges from prior work in that we fundamentally change the paragraph generation module to encourage topic consistency across sentences. 
%%{\color{blue}something along those lines has to go to the related work as well.}

As shown in Figure~\ref{fig:para_gen}, we encourage topic coherence across sentences by constructing \textit{Coherence Vectors}. In the following we describe the process of building these vectors. In order to compute the coherence vector for the $(i-1)^{\text{th}}$ sentence, we extract the hidden layer representation ($\in \mathbb{R}^H$) from the WordRNN, after having synthesized the last token of the $(i-1)^{\text{th}}$ sentence. %, post the generation of the final token of sentence $i$. 
This encoding carries information about  the $(i-1)^{\text{th}}$ sentence, and if favorably coupled with the topic vector $T_{i}$ of the $i^{\text{th}}$ sentence, encourages the theme of the $i^{\text{th}}$ sentence to be coherent with the previous one. However, for the aforementioned coupling to be successful, the hidden layer representation of the $(i-1)^{\text{th}}$ sentence still needs to be transformed to a representation that lies in the same space as the set of topic vectors. This transformation is achieved by propagating the final representation of the $(i-1)^{\text{th}}$ sentence through a 2-layer deep net of fully connected units, with the intermediate layer having $H$ activations. We used Scaled Exponential Linear Unit (SeLU) activations~\cite{klambauer2017self} for all neurons of this deep net. The output of this network is what we refer to as `Coherence Vector,' $C_{(i-1)}$.

\paragraph{Coupling Unit:}
\begin{figure}[t]
\centering
\includegraphics[scale=0.7]{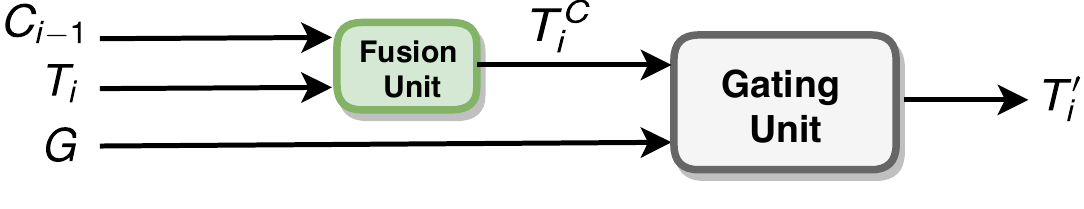}
\caption{The internal architecture of the `Coupling Unit'.} %{\color{blue}identical fontsize; `$C_{i-1}$' seems squashed - fixed}} 
\label{fig:couple}
\end{figure}

Having obtained the coherence vector $C_{i-1}$ from the $(i-1)^{\text{th}}$ sentence, a \textit{Coupling Unit} combines it with the topic vector of the next sentence, $T_i$, and the global topic representation $G$. This process is illustrated in Figure~\ref{fig:couple}.

More specifically, we first combine $C_{i-1}$ and $T_i$ into a vector  $T^C_{i} \in \mathbb{R}^H$ which is given by the solution to the following optimization problem:
\[
%\underset{T^C_{i+1}}{\text{min}}
T^C_{i} = \arg\min_{\hat T^C_{i}}
 \quad \alpha \lvert\lvert T_{i} - \hat T^C_{i} \rvert\rvert^2_2 + \beta \lvert\lvert C_{i-1} - \hat T^C_{i} \rvert\rvert^2_2 \quad \text{with}\quad \alpha, \beta \geq 0.
\]

The solution, when $\alpha, \beta$ both are not equal to 0, is given by: %, which is given by:
\[
T^C_{i} = \frac{\alpha T_{i} + \beta C_{i-1}}{\alpha + \beta}.
\]
We refer the interested reader to the supplementary for this derivation. 
Intuitively, this formulation encourages $T^C_{i}$ to be `similar' to both the coherence vector, $C_{i-1}$ and the current topic vector, $T_{i}$ -- thereby aiding cross-sentence topic coherence. Moreover, the closed form solution of this formulation permits an efficient implementation as well.
%{\color{blue}why do we use this approach rather than any other way to combine the two vectors?}

The obtained vector $T^C_{i}$ is then coupled with the global topic vector $G$, via a gating function. 
We implement this gating function using a single  GRU layer with vector $T^C_i$ as input and  global topic vector $G$ as its  hidden state vector. The output of this GRU cell, $T'_i$, is the final topic vector which is used to produce the $i^{\text{th}}$ sentence via the WordRNN. %Next, we present the loss function and the training formulation. %mechanism used for training our system.

%\paragraph{\textbf{Visual Representation:}} 
%To generate the topic vectors, we extract an efficient encoding of the input image, following Johnson \etal~\cite{johnson2016densecap}. More specifically, a cascaded application of a Convolutional Neural Network (CNN) along with a Region Proposal Network (RPN) gives fixed-length feature vectors for every detection of semantically salient regions in the image. The obtained set of vectors $\{v_1, \ldots, v_M\}$ with $v_i \in \mathbb{R}^D$ each correspond to a region in the image. We subsequently pool these vectors into a single  vector, $v \in \mathbb{R}^I$, using the approach suggested by Krause \etal~\cite{krause2017hierarchical}. {\color{blue}maybe some details here?} This pooled representation contains relevant information from the different semantically salient regions in the image, which is used as the basis for our topic generation module.

\paragraph{\textbf{Loss Function and Training:}}
%Our training data, consists of pairs $(x,y)$ of images and ground-truth paragraph. 
Both Topic Generation Net and Sentence Generation Net are trained jointly end-to-end using labeled training data, which consists of pairs $(x,y)$ of an image $x$ and a corresponding paragraph $y$. If one image is associated with multiple paragraphs, we create a separate pair for each.
Our training loss function $\ell_{\text{train}}(x,y)$ couples two cross-entropy losses: a binary cross-entropy sentence-level loss on the distribution $u_i$ for the $i^{th}$ sentence ($\ell_{\text{s}}(u_i, \mathbbm{1}_{i \leq S})$), and a word-level loss, on the  distribution $p_{i,j}$ for the $j^{th}$ word of the $i^{th}$ sentence ($\ell_{\text{w}}(p_{i,j}, y_{i,j})$). Assuming $S$ sentences in the ground-truth paragraph, with the $i^{th}$ sentence having $N_i$ words, our loss function is given by:
\begin{equation}
\ell_{\text{train}}(x,y) = \lambda_{s} \sum_{i=1}^S \ell_{\text{s}}(u_i, \mathbbm{1}_{i = S}) + \lambda_{w} \sum_{i=1}^S \sum_{j=1}^{N_i} \ell_{\text{w}}(p_{i,j}, y_{i,j}),
\label{eq:loss}
\end{equation}
where $\mathbbm{1}_{\{\cdot\}}$ is the indicator function, $\lambda_{s}, \lambda_{w}$ are the weights. Armed with this loss function our method is trained via the Adam optimizer~\cite{kingma2014adam} to update the parameters $w_T$ and $w_{s}$.

\subsection{Diverse Coherent Paragraph Generation}
\label{sec:Diverse}
\begin{figure}[t]
\centering
\includegraphics[scale=0.5]{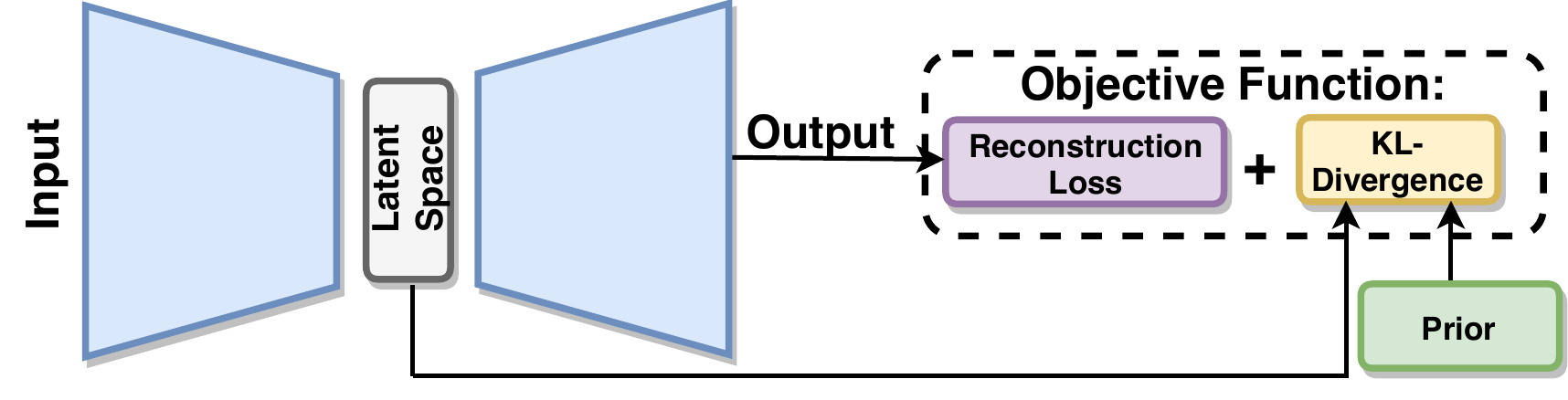}
\caption{General Framework of our VAE Formulation.} % {\color{blue}can we add a prior as an additional input to the KL-divergence - done}} 
\label{fig:gen_vae}
\end{figure}

The aforementioned scheme for generating paragraphs lacks in one key aspect: it doesn't model the ambiguity inherent to a \textit{diverse} set of paragraphs that fit a given image. In order to incorporate this element of diversity into our model, we cast the designed paragraph generation mechanism into a \textit{Variational Autoencoder} (VAE)~\cite{kingma2013auto} formulation, a generic architecture of which is shown in Figure~\ref{fig:gen_vae}. Note that we prefer a VAE formulation over other popular tools for modeling diversity, such as GANs, because of the following reasons: (1) GANs are known to suffer from training difficulties unlike VAEs~\cite{arjovsky2017wasserstein}; (2) The intermediate sampling step in the generator of a GAN (for generating text) is not differentiable and thus one has to resort to Policy Gradient-based algorithms or Gumbel softmax, which makes the training procedure non-trivial. The details of our formulation follow. %in greater detail.

\paragraph{\textbf{VAE Formulation:}} 
%{\color{red}new notation is introduced here, \eg, $x$ hasn't been discussed before; this is not adequate; we should follow notation which was introduced earlier; since words \etc weren't introduced before we probably need to do that; we should probably do that in any case.}

%{\color{blue}please describe this more carefully; right now many statements are plain wrong; check the papers that I mentioned}

%A VAE consists of an encoder and a decoder coupled via a latent space representation (see Figure~\ref{fig:gen_vae}). Let $z$ denote the embedding of an input paragraph $y$, conditioned on an image $x$. 
The goal of our VAE formulation is to model the log-likelihood of  paragraphs $y$ conditioned on images $x$, \ie,  $\ln p(y|x)$. To this end, a VAE assumes that the data, \ie, in our case  paragraphs, arise from a low-dimensional manifold space represented by samples $z$. %, which is parameterized by the latent embeddings, $z$. 
%<<<<<<< HEAD
Given a sample $z$, we reconstruct, \ie, decode, a paragraph $y$ by modeling $p_\theta(y|z, x)$ via a deep net. The ability to randomly sample from this latent space provides  diversity. In the context of our task the decoder is the paragraph generation module described in \secref{sec:Coherence}, augmented by taking samples from the latent space as input. We subsequently denote the parameters of the paragraph generation module by $\theta = [w_T, w_\text{s}]$. %Computing this distribution, $p_\theta(y|z, x)$, %requires computing the posterior of the distribution $p_\theta(z|y, x)$. {\color{blue}the preceding statement is wrong; this is not required but the posterior is needed for other reasons; please check the papers which I mentioned.}
%=======
%Given a sample $z$, we reconstruct a paragraph $y$ by modeling $p_\theta(y|z, x)$ via a deep net. In the context of our task the decoder is the paragraph generation module described in Sec \ref{sec:Coherence}. We subsequently denote the parameters of the paragraph generation module by $\theta = [w_T, w_\text{s}]$.%, w_X]$. %Computing this distribution, $p_\theta(y|z, x)$, %requires computing the posterior of the distribution $p_\theta(z|y, x)$. {\color{blue}the preceding statement is wrong; this is not required but the posterior is needed for other reasons; please check the papers which I mentioned.}
%>>>>>>> c8bd1c9bb2cc283de52ecdd27f9b0ce363603989
To learn a meaningful manifold space we require the decoder's posterior $p_\theta(z|y, x)$. 
However computing the decoder's posterior $p_\theta(z|y, x)$ is known to be challenging~\cite{kingma2013auto}. Hence, we commonly approximate this distribution using another probability $q_\phi(z|y, x)$, which constitutes the encoder section of the model, parameterized by $\phi$. Further, let $p(z)$ denote the prior distribution of samples in the latent space. Using the aforementioned distributions, the VAE formulation can be obtained from the following identity: %{\color{blue}check the papers; this is not the VAE objective}
%A VAE consists of an encoder, and a decoder coupled via a latent space representation (see Figure~\ref{fig:gen_vae}). Let $z$ denote the embedding of an input paragraph $y$, conditioned on an image $x$. The goal here, at training time, is to reconstruct a close enough approximation of $y$. However, different from stacked autoencoders~\cite{vincent2010stacked}, besides ensuring a close-enough reconstruction, VAEs impose a smoothness constraint between the conditional distribution of sample encodings, $z$, given the input pair $(x,y)$ - $q_\phi(z|x,y)$ and the distribution over the samples in the latent space ($p_\theta(z)$). This renders the latent space to be a manifold. The training is performed by maximizing the log-likelihood of the data~\cite{kingma2013auto}:
%the here, at training time, is to reconstruct a close enough approximation of $y$. However, different from stacked autoencoders~\cite{vincent2010stacked}, besides ensuring a close-enough reconstruction, VAEs impose a smoothness constraint between the conditional distribution of sample encodings, $z$, given the input pair $(x,y)$ - $q_\phi(z|x,y)$ and the distribution over the samples in the latent space ($p_\theta(z)$). This renders the latent space to be a manifold. The training is performed by maximizing the log-likelihood of the data~\cite{kingma2013auto}:

% \underset{z}{\Sigma} q_\phi(z|x,y) \ln p_\theta(y) = \underbrace{\Sigma_z q_\phi(z|x,y) \ln \frac{p_\theta(y, z|x)}{q_\phi(z|y, x)}}_{\mathcal{L}(q_\phi(z|y,x), p_\theta(y, z|x))} + KL(q_\phi(z|y,x), p_\theta(z|y,x))
\[
\ln p(y|x)\! -\! \text{KL}(q_\phi(z|y, x), p_\theta(z|y, x)) \!=\! \mathbb{E}_{q_\phi(z|y, x)} [\ln p_\theta(y|z, x)] \!-\! \text{KL}(q_\phi(z|y, x), p(z)),
\]
%More specifically, assuming that the encoder, $q_\phi(z|y, x)$, has sufficient capacity to mimic the posterior distribution of the decoder, $p_\theta(z|y, x)$, \ie, the two are a close approximation of each other, we see that maximizing the log-likelhood amounts to maximizing the right-hand side of the above equation, with respect to the encoder and decoder parameters $\phi$ and $\theta$ respectively. Thus the expression on the right hand side of the equation constitutes the VAE objective. 
 where $\text{KL}(\cdot, \cdot)$ denotes the KL divergence between two distributions. Due to the non-negativity of the KL-divergence we immediately observe the right hand side to be a lower bound on the log-likelihood $\ln p(y|x)$ which can be maximized \wrt its parameters $\phi$ and $\theta$. 
The first term on the right hand side optimizes the reconstruction loss, \ie, the conditional likelihood of the decoded paragraph (for which we use the loss in Equation~\ref{eq:loss}), while the second term acts like a distributional regularizer (ensuring smoothness). %The first term is equivalent to optimizing the loss in Equation~\ref{eq:loss}.
Training this system end-to-end via backpropagation is hard because of the intermediate, non-differentiable, step of sampling $z$. This bottleneck is mitigated by introducing the \textit{Re-parameterization Trick}~\cite{kingma2013auto}. The details of the encoder and decoder follow. %sections of our network in more detail.
%At test time, one randomly samples a $z$ from the distribution underlying the latent space and passes it thorough the decoder to generate a sample. This brings in the element of diversity.

\begin{figure}[t]
\centering
\includegraphics[width=\textwidth]{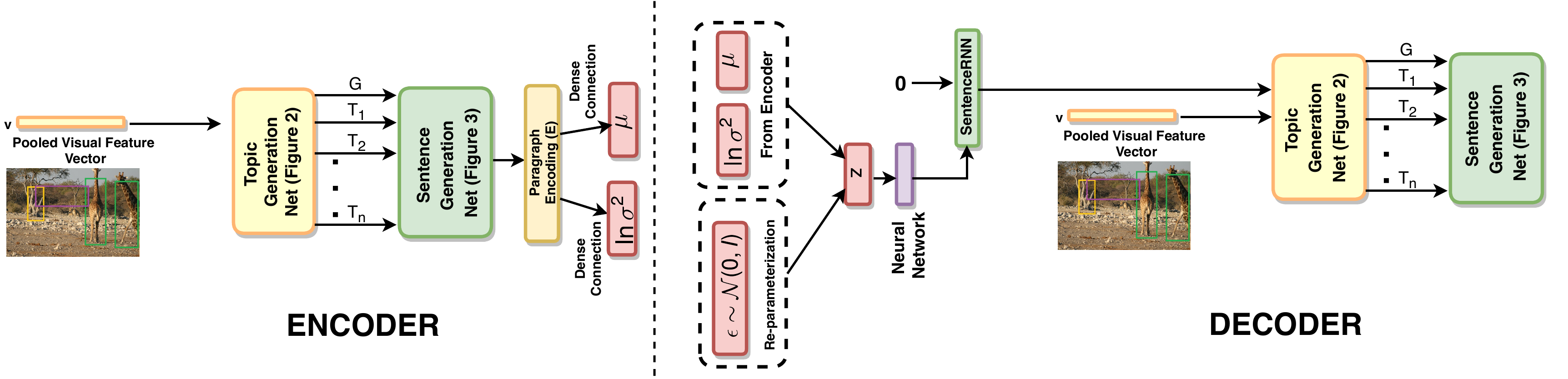}
\caption{Architecture of the Encoder and Decoder of our VAE formulation.}
\label{fig:enc_dec}
%\centering
%\includegraphics[height=3cm,width=8cm]{New_figs/VAE_Dec.png}
%\caption{Architecture of the Decoder Module of our model in the VAE setting.}
%\label{fig:dec}
\end{figure}

\paragraph{\textbf{Encoder:}} The encoder  architecture is shown in Figure~\ref{fig:enc_dec}. %It extends our original models shown in Figures~\ref{fig:topic_gen}, \ref{fig:para_gen} by {\color{blue}someting}. 
Given the image $x$ and a ground-truth paragraph $y$ we encode the sample $(x,y)$ by passing it through the topic and sentence generation nets. We then extract the hidden state vector ($E \in \mathbb{R}^H$) from the final WordRNN of the Sentence Generation net. %, post the generation of the last token of the paragraph. 
This vector is passed through a 1-layer densely connected net, the output layer of which has $2H$ neurons. We assume the conditional distribution underlying the encoder, $q_\phi(z|y,x)$ to be a Gaussian, whose mean $\mu$ is the output of the first $H$ neurons, while the remaining $H$ neurons give a measure of the log-variance, \ie, $\ln\sigma^2$.
%  of the distribution
%\begin{figure}[t]
%\end{figure}

\paragraph{\textbf{Decoder:}} The decoding architecture is also shown in Figure~\ref{fig:enc_dec}. While decoding, we draw a sample $z \sim \mathcal{N}(0,I)$ ($z \in \mathbb{R}^H$, for training: we additionally shift and scale it by: $z = \mu + \sigma \epsilon$, where $\epsilon \sim \mathcal{N}(0,I)$) and pass it to the SentenceRNN, via a single-layer neural net with $I$ output neurons. The hidden state of this RNN is then forward propagated to the SentenceRNN unit, which also receives the pooled visual vector $v$. Afterwards, the decoding proceeds as discussed before.

%In the next section, we proceed to discuss the details of our empirical evaluation.  %In the following we discuss, our findings from empirical evaluations.
%During the training, the output sentences from the Word RNN allow us to compute the word-level loss, while the Sentence RNN gives us a measure of loss based on the difference of the number of sentences in the generated paragraph versus that in the ground truth~\cite{krause2017hierarchical}. These loses are backpropagated all the way through both the decoder and encoder in order to update the model parameters.

%% file: evaluation.tex
% !TEX root = main.tex
\section{Experimental Evaluations}
\label{sec:exp}
\begin{figure}[t]
 %\centering
 \includegraphics[scale=0.25,center]{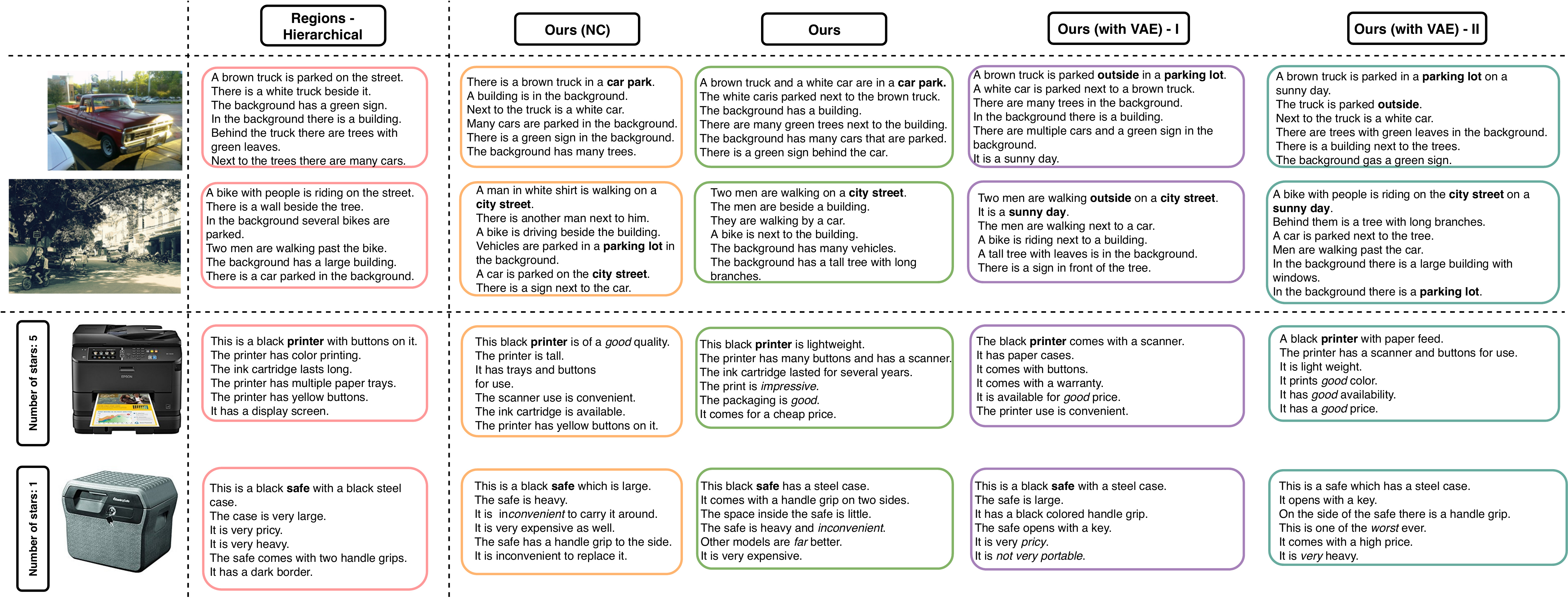} % [width=\textwidth]
 \caption{Paragraphs generated under different settings with our developed approach, vis-\`a-vis Regions-Hierarchical~\cite{krause2017hierarchical}. The first, and second images are from the Stanford dataset, while the third and fourth images are from the Amazon dataset.} % {\color{blue} don't change the aspect ratio of the figure since the text would look squeezed otherwise which is unprofessional - fixed; add two (better three) more rows for each of the datasets - not enough space.}} %competing methods.}
 \label{fig:samp}
 \end{figure}

%In order to verify the efficacy of our  approach, we compare our algorithm to several competing baselines on two datasets.
\paragraph{\textbf{Datasets:}} We first conduct experiments on the \textit{Stanford image-paragraph dataset}~\cite{krause2017hierarchical}, a standard in the area of visual paragraph generation. The dataset consists of 19,551 images from the Visual Genome~\cite{krishna2017visual} and MS COCO dataset~\cite{lin2014microsoft}. These images are annotated with human-labeled paragraphs, 67.50 words long, with each sentence having 11.91 words, on average. The experimental protocol divides this dataset into 14,575 training, 2,487 validation, and 2,489 testing examples~\cite{krause2017hierarchical}. Further, in order to exhibit generalizability of our  approach, different from prior work, we also undertake experiments on the much larger, \textit{Amazon Product-Review dataset} (`Office-Products' category)~\cite{mcauley2015image} for the task of generating reviews. This is a dataset of images of common categories of office-products, such as printer, pens, \etc (see Figure~\ref{fig:samp}), crawled from amazon.com. There are 129,970 objects in total, each of which belongs to a category of office products. For every object, there is an associated image, captured in an uncluttered setting with sufficient illumination. Accompanying the image, are multiple reviews by users of the product. Further, each review is supplemented by a star rating, an integer between 1 (poor) and 5 (good). On an average there are 6.4 reviews per star rating per object. A review is 71.66 words long, with 13.52 words per sentence, on average. We randomly divide the dataset into 5,000 test, and 5,000 validation examples, while the remaining examples are used for training.

\paragraph{\textbf{Baselines:}} We compare our approach to several recently introduced and our own custom designed baselines. Given an image, `Image-Flat' directly synthesizes a paragraph, token-by-token, via a single RNN~\cite{karpathy2015deep}. `Regions-Hierarchical' on the other hand, generates a paragraph, sentence by sentence~\cite{krause2017hierarchical}. Liang \etal~\cite{liang2017recurrent} essentially train the approach of Krause \etal~\cite{krause2017hierarchical} in a GAN setting (`RTT-GAN'), coupled with an attention mechanism. However, Liang \etal also report results on the Stanford image-paragraph dataset by using additional training data from the MS COCO dataset, which we refer to as `RTT-GAN (Plus).' We also train our model in a GAN setting and indicate this baseline as `Ours (GAN).' Additionally, we create baselines for our model without coherence vectors, essentially replacing them with a zero vector for every time-step. We refer to this baseline as `Ours (NC).' In another setting, we only set the global topic vector to zero for every time-step. We refer to this baseline as `Ours (NG).' %The last two baselines also serve as an ablation analysis.

\paragraph{\textbf{Evaluation Metrics:}} We report the performance of all models on 6 widely used language generation metrics: BLEU-\{1, 2, 3, 4\}~\cite{papineni2002bleu}, METEOR~\cite{denkowski2014meteor}, and CIDEr~\cite{vedantam2015cider}. While the BLEU scores largely measure just the n-gram precision, METEOR, and CIDEr  are known to provide a more robust evaluation of language generation algorithms~\cite{vedantam2015cider}.
%  (which measures F-measure of precision and recall) (which measures cosine similarity between the candidate and ground-truth)
\paragraph{\textbf{Implementation Details:}} For the Stanford dataset, we set the dimension of the pooled visual feature vector, $v$, to be 1024. For the Amazon dataset, however, we use a visual representation obtained from VGG-16~\cite{simonyan2014very}. Since, these images are generally taken with just the principal object in view (see Figure~\ref{fig:samp}), a standard CNN suffices. We extract representations from the penultimate fully connected layer of the CNN, giving us a vector of 4,096 dimensions. Hence, we use a single-layer neural network to map this vector to the input vector of 1,024 dimensions. For both SentenceRNN and WordRNN, the GRUs have hidden layers ($H$) of 512 dimension. For the Amazon dataset, we condition the first SentenceRNN, with an $H$-dimensional embedding of the number of stars. We set $\lambda_s, \lambda_w$ to be 5.0, and 1.0 respectively, the maximum number of sentences per paragraph, $S_{\text{max}}$, to be 6, while the maximum number of words per sentence is set to be 30, for both datasets. In the coupling unit, $\alpha$ is set to 1.0, and $\beta$ is set to 1.5 for the Stanford dataset, while for the Amazon dataset the corresponding values are 1.0 and 3.0. The learning rate of the model is set to 0.0001 for the first 5 epochs and is halved every 5 epochs after that, for both datasets. These hyper-parameters are chosen by optimizing the performance, based on the average of METEOR and CIDEr scores, on the validation set for both datasets. We use the same vocabulary as Krause~\etal~\cite{krause2017hierarchical}, for the Stanford dataset, while a vocabulary size of the $11,000$ most frequent words is used for the Amazon dataset. Additional implementational details can be found on the project website\footnote{\url{https://sites.google.com/site/metrosmiles/research/research-projects/capg_revg}}. For purposes of comparison, for the Amazon dataset, we run our implementation of all baselines, with their hyper-parameters picked based on a similar protocol,  while for the Stanford dataset we report performance for prior approaches directly from~\cite{liang2017recurrent}. %Additionally for Regions-Hierarchical~\cite{krause2017hierarchical}, we report the performance of our implementation of the model, separately. We denote this implementation as `Reg-Hier. (Our Ver)'.
%{\color{blue}did we make sure to reproduce the numbers when the stanford dataset before running on amazon? if yes, we should mention - done}
% In our VAE-decoder, we first map the $z$ to a 1,024 dimension space prior to passing it as an input to the sentence RNN. This mapping is achieved via a single-layer neural network.
\paragraph{\textbf{Results:}} 
% Include the results paragraph here
\begin{table}[t]
  \centering
  \caption{Comparison of captioning performance on the Stanford Dataset}% of various competing algorithms.}
  \begin{tabular}{l||c|c|c|c||c||c}
    \hline \hline
    \textbf{Method} & \textbf{BLEU-1} & \textbf{BLEU-2} & \textbf{BLEU-3}    & \textbf{BLEU-4}    &  \textbf{METEOR}   & \textbf{CIDEr}  \\ \hline
    Image-Flat~\cite{karpathy2015deep} & 34.04 & 19.95 & 12.2 & 7.71 & 12.82 & 11.06 \\
    Regions-Hierarchical~\cite{krause2017hierarchical} & 41.9 & 24.11 & 14.23 & 8.69 & 15.95 & 13.52 \\
    %Reg-Hier. (Our Ver)~\cite{krause2017hierarchical} & 41.83 & 24.14 & 14.21 & 8.65 & 15.87 & 13.48 \\
    RTT-GAN~\cite{liang2017recurrent} & 41.99 & 24.86 & 14.89 & 9.03 & 17.12 & 16.87  \\
    RTT-GAN (Plus)~\cite{liang2017recurrent}   & 42.06 & 25.35 & 14.92 & 9.21 & 18.39 & 20.36  \\ \hline
        Ours (NC)   & 42.03 & 24.84 & 14.47 & 8.82 & 16.89 & 16.42 \\
        Ours (NG)   & 42.05 & 25.05 & 14.59 & 8.96 & 17.26 & 18.23 \\
        Ours    & 42.12 & 25.18 & 14.74 & 9.05 & 17.81 & 19.95 \\
        Ours (with GAN)   & 42.04 & 24.96 & 14.53 & 8.95 & 17.21 & 18.05 \\
        Ours (with VAE)    & \textbf{42.38} & \textbf{25.52} & \textbf{15.15} & \textbf{9.43} & \textbf{18.62} & \textbf{20.93}\\
    \hline 
    Human (as in \cite{krause2017hierarchical})   & 42.88 & 25.68 & 15.55 & 9.66 & 19.22 & 28.55\\
    \hline \hline
  \end{tabular}
  \label{tab:stanford}
\end{table}
\begin{table} [t]
  \centering
  \caption{Comparison of captioning performance on the Amazon Dataset}% of various competing algorithms.}
  \begin{tabular}{l||c|c|c|c||c||c}
    \hline \hline
    \textbf{Method} & \textbf{BLEU-1} & \textbf{BLEU-2} & \textbf{BLEU-3}    & \textbf{BLEU-4}    &  \textbf{METEOR}   & \textbf{CIDEr}  \\ \hline
    Image-Flat~\cite{karpathy2015deep} & 40.31 & 30.63 & 25.32 & 15.64 & 10.97 & 9.63 \\
    Regions-Hierarchical~\cite{krause2017hierarchical} & 45.74 & 34.8 & 27.54 & 16.67 & 14.23 & 12.02 \\
    RTT-GAN~\cite{liang2017recurrent} & 45.93 & 36.42 & 28.28 & 17.26 & 16.29 & 15.67 \\ \hline
        Ours (NC)   & 45.85 & 35.97 & 27.96 & 16.98 & 15.86 & 15.39 \\
        Ours (NG)   & 45.88 & 36.33 & 28.15 & 17.17 & 16.04 & 15.54 \\
        Ours    & 46.01 & 36.86 & 28.73 & 17.45 & 16.58 & 16.05 \\
        Ours (with GAN)   & 45.86 & 36.25 & 28.07 & 17.06 & 15.98 & 15.43 \\
        Ours (with VAE)    & \textbf{46.32} & \textbf{37.45} & \textbf{29.42} & \textbf{18.01} & \textbf{17.64} & \textbf{17.17}\\
    \hline \hline
  \end{tabular}
  \label{tab:Amazon}
\end{table}

%evaluate our method as well as all baselines on the less common {\color{blue}xxx} dataset~\cite{},  
%{\color{blue}xxx}~\cite{}. 
%In order to avoid overfitting to a single dataset we also evaluate our method as well as all baselines on the less common {\color{blue}xxx} dataset~\cite{}, showing {\color{blue}xxx}. We conclude by showing qualitative results and an ablation study of the developed approach.

% , demonstrating state-of-the-art performance
%\paragraph{xxx dataset:} A standard dataset for visual paragraph generation.

%\noindent{\bf Dataset:} The {\color{blue}xxx} dataset consists of {\color{blue}description of dataset and properties goes here.}

%\noindent{\bf Evaluation metrics:} {\color{blue}discuss evaluation metrics here}

%\noindent{\bf Results:}
Tables~\ref{tab:stanford} and~\ref{tab:Amazon} show the performance of our algorithm vis-\`a-vis other comparable baselines. Our model, especially when trained in the VAE setting, outperforms all other baselines (on all 6 metrics). %, nearing human-level captioning performance on the Stanford dataset. 
Even the models trained under the regular (non-VAE) setup  outperform most of the baselines and are comparable to the approach of Liang \etal~\cite{liang2017recurrent}, an existing state-of-the-art for this task. Our performance on the rigorous METEOR and CIDEr metrics, on both datasets, attest to our improved paragraph generation capability. The capacity to generate diverse paragraphs, using our VAE setup, pays off especially well on the Amazon dataset, since multiple reviews with the same star rating are associated with an object, creating an inherent ambiguity. %in the mapping of input images and the number of stars to the output review. 
Noticeably, our model is worse off in terms of performance, when trained under the GAN setting. This observation is along the lines of prior work~\cite{dai2017towards}. We surmise that this results from the  difficulty of training GANs~\cite{arjovsky2017wasserstein} in conjunction with the fact that the GAN-based setup isn't trained directly with maximum-likelihood.

\paragraph{Qualitative results:} Figure~\ref{fig:samp} presents a sampling of our generated paragraphs. The first example in the figure (the first row) shows that our model can generate coherent paragraphs, while capturing meta-concepts like `car-park' or `parking lot,' from images with complex scenes. Regions-Hierarchical~\cite{krause2017hierarchical} faces challenges to incorporate these `meta-concepts' into the generated paragraphs. For several of the instances in the Amazon dataset (such as the images in the third and fourth rows), both our method and Regions-Hierarchical~\cite{krause2017hierarchical} successfully detect the principal object in the image.  We speculate that this is due to easy object recognition for images of the Amazon dataset, and to a lesser extent due to an improved paragraph generation algorithm.  Additionally in the VAE setting, we are able to generate two distinctly different paragraphs with the same set of inputs, just by sampling a different $z$ each time (the two rightmost columns in Figure~\ref{fig:samp}), permitting our results to be diverse. Moreover, for the Amazon dataset (third and fourth rows in Figure~\ref{fig:samp}) we see that our model learns to synthesize `sentiment' words depending on the number of input stars. We present additional visualizations in the supplementary material.

%\paragraph{xxx dataset:} A dataset from Amazon.

%\noindent{\bf Dataset:} The {\color{blue}xxx} dataset consists of {\color{blue}description of dataset and properties goes here.}

%\noindent{\bf Evaluation metrics:} {\color{blue}discuss evaluation metrics here; same ones as before?}

%\noindent{\bf Results:}

%\paragraph{Qualitative results:}

\paragraph{Ablation study:} %We conduct ablation analysis to judge the importance of coherence vectors.
In one setting, we judge the importance of coherence vectors, by just using the global vector and setting the coherence vectors to 0, in the sentence generation net. The results for this setting (`Ours (NC)') are shown in Tables~\ref{tab:stanford} and \ref{tab:Amazon}, while qualitative results are shown in Figure~\ref{fig:samp}. These numbers reveal that just by incorporating the global topic vector it is feasible to generate reasonably good paragraphs. However, incorporating coherence vectors makes the synthesized paragraphs more human-like. A look at the second column of Figure~\ref{fig:samp} shows that even without coherence vectors we are able to detect the central underlying image theme, like `car-park' but the sentences seem to exhibit sharp topic transition, quite like the Regions-Hierarchical~\cite{krause2017hierarchical} approach. We  rectify this by introducing coherence vectors. %, when coherent vectors are introduced. 

In another setting, we set the global topic vector to 0, at every time-step, while retaining the coherence vectors. The performance in this setting is indicated by `Ours (NG)' in Tables~\ref{tab:stanford} and \ref{tab:Amazon}. The results suggest that incorporating the coherence vectors is much more critical for improved paragraph generation. %, as compared to the global topic vector.

%% file: conclusion.tex
% !TEX root = main.tex
\section{Conclusions and Future Work}
\label{sec:conc}
%We developed `consistency vectors' which explicitly ensure consistency between generated sentences for paragraph generation. We demonstrated efficacy of the proposed technique on the standard dataset for paragraph generation and an additional dataset for generation of reviews, showing that `consistency vector' based methods help to achieve state-of-the-art performance on both. In the future we plan to extend our technique to generation of stories, ensuring a hierarchical flow of information.

In this work, we developed `coherence vectors' which explicitly ensure consistency of themes between generated sentences during  paragraph generation. Additionally, the `global topic vector' was designed to capture the underlying main plot of an image. We demonstrated the efficacy of the proposed technique on two datasets, showing that our model when trained with effective autoencoding techniques can achieve state-of-the-art performance for both caption and review generation tasks. In the future we plan to extend our technique for the task of generation of even longer narratives, such as stories. %, which need to ensure a hierarchical flow of information. Another promising direction to refine the paragraph generation capability of existing models is the combination with part-of-speech structure. % of  different sentences of a paragraph, in order .

\noindent\textbf{Acknowledgments:} This material is based upon work supported in part by the National Science Foundation under Grant No.~1718221, Samsung, and 3M. We thank NVIDIA for providing the GPUs used for this research. MC acknowledges Prof. Narendra Ahuja for providing logistical support for this research.

%% file: arxiv_main.bbl
\begin{thebibliography}{10}
\providecommand{\url}[1]{\texttt{#1}}
\providecommand{\urlprefix}{URL }
\providecommand{\doi}[1]{https://doi.org/#1}

\bibitem{AnejaCVPR2018}
Aneja, J., Deshpande, A., Schwing, A.G.: {Convolutional Image Captioning}. In:
  Proc. CVPR (2018)

\bibitem{antol2015vqa}
Antol, S., Agrawal, A., Lu, J., Mitchell, M., Batra, D., Lawrence~Zitnick, C.,
  Parikh, D.: Vqa: Visual question answering. In: Proc. ICCV (2015)

\bibitem{arjovsky2017wasserstein}
Arjovsky, M., Chintala, S., Bottou, L.: Wasserstein gan. arXiv preprint 2017

\bibitem{chatterjee2015novel}
Chatterjee, M., Leuski, A.: A novel statistical approach for image and video
  retrieval and its adaption for active learning. In: Proc. ACM Multimedia
  (2015)

\bibitem{chen2015mind}
Chen, X., Lawrence~Zitnick, C.: Mind's eye: A recurrent visual representation
  for image caption generation. In: Proc. CVPR (2015)

\bibitem{chung2014empirical}
Chung, J., Gulcehre, C., Cho, K., Bengio, Y.: Empirical evaluation of gated
  recurrent neural networks on sequence modeling. arXiv preprint 2014

\bibitem{chung2015recurrent}
Chung, J., Kastner, K., Dinh, L., Goel, K., Courville, A.C., Bengio, Y.: A
  recurrent latent variable model for sequential data. In: Proc. NIPS (2015)

\bibitem{dai2017towards}
Dai, B., Fidler, S., Urtasun, R., Lin, D.: Towards diverse and natural image
  descriptions via a conditional gan. arXiv preprint 2017

\bibitem{denkowski2014meteor}
Denkowski, M., Lavie, A.: Meteor universal: Language specific translation
  evaluation for any target language. In: Proc. ninth workshop on statistical
  machine translation (2014)

\bibitem{DeshpandeARXIV2018}
Deshpande, A., Aneja, J., Wang, L., Schwing, A.G., Forsyth, D.A.: {Diverse and
  Controllable Image Captioning with Part-of-Speech Guidance}. In:
  https://arxiv.org/abs/1805.12589 (2018)

\bibitem{donahue2015long}
Donahue, J., Anne~Hendricks, L., Guadarrama, S., Rohrbach, M., Venugopalan, S.,
  Saenko, K., Darrell, T.: Long-term recurrent convolutional networks for
  visual recognition and description. In: Proc. CVPR (2015)

\bibitem{gao2015you}
Gao, H., Mao, J., Zhou, J., Huang, Z., Wang, L., Xu, W.: Are you talking to a
  machine? dataset and methods for multilingual image question. In: Proc. NIPS
  (2015)

\bibitem{goodfellow2014generative}
Goodfellow, I., Pouget-Abadie, J., Mirza, M., Xu, B., Warde-Farley, D., Ozair,
  S., Courville, A., Bengio, Y.: Generative adversarial nets. In: Proc. NIPS
  (2014)

\bibitem{gregor2015draw}
Gregor, K., Danihelka, I., Graves, A., Rezende, D.J., Wierstra, D.: Draw: A
  recurrent neural network for image generation. arXiv preprint 2015

\bibitem{JainCVPR2018}
Jain, U., Lazebnik, S., Schwing, A.G.: {Two can play this Game: Visual Dialog
  with Discriminative Question Generation and Answering}. In: Proc. CVPR (2018)

\bibitem{jain2017creativity}
Jain, U., Zhang, Z., Schwing, A.: Creativity: Generating diverse questions
  using variational autoencoders. In: Proc. CVPR (2017)

\bibitem{johnson2016densecap}
Johnson, J., Karpathy, A., Fei-Fei, L.: Densecap: Fully convolutional
  localization networks for dense captioning. In: Proc. CVPR (2016)

\bibitem{karpathy2015deep}
Karpathy, A., Fei-Fei, L.: Deep visual-semantic alignments for generating image
  descriptions. In: Proc. CVPR (2015)

\bibitem{kingma2014adam}
Kingma, D.P., Ba, J.: Adam: A method for stochastic optimization. arXiv
  preprint 2014

\bibitem{kingma2013auto}
Kingma, D.P., Welling, M.: Auto-encoding variational bayes. arXiv preprint 2013

\bibitem{klambauer2017self}
Klambauer, G., Unterthiner, T., Mayr, A., Hochreiter, S.: Self-normalizing
  neural networks. In: Proc. NIPS (2017)

\bibitem{krause2017hierarchical}
Krause, J., Johnson, J., Krishna, R., Fei-Fei, L.: A hierarchical approach for
  generating descriptive image paragraphs. In: Proc. CVPR (2017)

\bibitem{krishna2017visual}
Krishna, R., Zhu, Y., Groth, O., Johnson, J., Hata, K., Kravitz, J., Chen, S.,
  Kalantidis, Y., Li, L.J., Shamma, D.A., Bernstein, M.S., Fei-Fei, L.: Visual
  genome: Connecting language and vision using crowdsourced dense image
  annotations. IJCV  (2017)

\bibitem{krizhevsky2012imagenet}
Krizhevsky, A., Sutskever, I., Hinton, G.E.: Imagenet classification with deep
  convolutional neural networks. In: Proc. NIPS (2012)

\bibitem{lavrenko2004model}
Lavrenko, V., Manmatha, R., Jeon, J.: A model for learning the semantics of
  pictures. In: Proc. NIPS (2004)

\bibitem{liang2017recurrent}
Liang, X., Hu, Z., Zhang, H., Gan, C., Xing, E.P.: {Recurrent Topic-Transition
  GAN for Visual Paragraph Generation}. In: Proc. ICCV (2017)

\bibitem{lin2014microsoft}
Lin, T.Y., Maire, M., Belongie, S., Hays, J., Perona, P., Ramanan, D.,
  Doll{\'a}r, P., Zitnick, C.L.: Microsoft coco: Common objects in context. In:
  Proc. ECCV (2014)

\bibitem{malinowski2015ask}
Malinowski, M., Rohrbach, M., Fritz, M.: Ask your neurons: A neural-based
  approach to answering questions about images. In: Proc. ICCV (2015)

\bibitem{mao2014deep}
Mao, J., Xu, W., Yang, Y., Wang, J., Huang, Z., Yuille, A.: Deep captioning
  with multimodal recurrent neural networks (m-rnn). arXiv preprint 2014

\bibitem{mcauley2015image}
McAuley, J., Targett, C., Shi, Q., Van Den~Hengel, A.: Image-based
  recommendations on styles and substitutes. In: Proc. ACM SIGIR (2015)

\bibitem{pan2004automatic}
Pan, J.Y., Yang, H.J., Duygulu, P., Faloutsos, C.: Automatic image captioning.
  In: Proc. ICME (2004)

\bibitem{papineni2002bleu}
Papineni, K., Roukos, S., Ward, T., Zhu, W.J.: Bleu: a method for automatic
  evaluation of machine translation. In: Proc. ACL (2002)

\bibitem{ren2015exploring}
Ren, M., Kiros, R., Zemel, R.: Exploring models and data for image question
  answering. In: Proc. NIPS (2015)

\bibitem{SchwartzNIPS2017}
Schwartz, I., Schwing, A.G., Hazan, T.: {High-Order Attention Models for Visual
  Question Answering}. In: Proc. NIPS (2017)

\bibitem{shih2016look}
Shih, K.J., Singh, S., Hoiem, D.: Where to look: Focus regions for visual
  question answering. In: Proc. CVPR (2016)

\bibitem{simonyan2014very}
Simonyan, K., Zisserman, A.: Very deep convolutional networks for large-scale
  image recognition. arXiv preprint 2014

\bibitem{vedantam2015cider}
Vedantam, R., Lawrence~Zitnick, C., Parikh, D.: Cider: Consensus-based image
  description evaluation. In: Proc. CVPR (2015)

\bibitem{vinyals2015show}
Vinyals, O., Toshev, A., Bengio, S., Erhan, D.: Show and tell: A neural image
  caption generator. In: Proc. CVPR (2015)

\bibitem{WangNIPS2017}
Wang, L., Schwing, A.G., Lazebnik, S.: {Diverse and Accurate Image Description
  Using a Variational Auto-Encoder with an Additive Gaussian Encoding Space}.
  In: Proc. NIPS (2017)

\bibitem{xiao2007fusion}
Xiao, Y., Chua, T.S., Lee, C.H.: Fusion of region and image-based techniques
  for automatic image annotation. In: Proc. International Conference on
  Multimedia Modeling (2007)

\bibitem{xie2018thesis}
Xie, P.: {Diversity-Promoting and Large-Scale Machine Learning for Healthcare}.
  \url{http://www.cs.cmu.edu/~pengtaox/thesis_proposal_pengtaoxie.pdf} (2018),
  [Online; accessed 25-July-2018]

\bibitem{xiong2016dynamic}
Xiong, C., Merity, S., Socher, R.: Dynamic memory networks for visual and
  textual question answering. In: Proc. ICML (2016)

\bibitem{xu2016ask}
Xu, H., Saenko, K.: Ask, attend and answer: Exploring question-guided spatial
  attention for visual question answering. In: Proc. ECCV (2016)

\bibitem{xu2015show}
Xu, K., Ba, J., Kiros, R., Cho, K., Courville, A., Salakhudinov, R., Zemel, R.,
  Bengio, Y.: Show, attend and tell: Neural image caption generation with
  visual attention. In: Proc. ICML (2015)

\bibitem{yang2016stacked}
Yang, Z., He, X., Gao, J., Deng, L., Smola, A.: Stacked attention networks for
  image question answering. In: Proc. CVPR (2016)

\bibitem{you2016image}
You, Q., Jin, H., Wang, Z., Fang, C., Luo, J.: Image captioning with semantic
  attention. In: Proc. CVPR (2016)

\bibitem{yu2016video}
Yu, H., Wang, J., Huang, Z., Yang, Y., Xu, W.: Video paragraph captioning using
  hierarchical recurrent neural networks. In: Proc. CVPR (2016)

\end{thebibliography}
